%% file: main.tex
\documentclass{article}

\PassOptionsToPackage{numbers, compress}{natbib}

\usepackage[preprint]{neurips_2025}

\usepackage[utf8]{inputenc} 
\usepackage[T1]{fontenc}    
\usepackage[colorlinks]{hyperref}       
\usepackage{url}            
\usepackage{booktabs}       
\usepackage{amsfonts}       
\usepackage{nicefrac}       
\usepackage{microtype}      
\usepackage{xcolor}         
\usepackage{xspace}
\usepackage{lipsum}
\usepackage{colortbl}
\usepackage{graphicx}
\usepackage{amsmath}
\usepackage{multirow}
\usepackage{wrapfig}
\usepackage{subcaption}
\usepackage{amssymb}
\usepackage{pifont}
\usepackage{enumitem}

\usepackage{soul}  
\sethlcolor{blue}  

\definecolor{delectricblue}{RGB}{93, 117, 210}
\colorlet{lightdelectricblue}{delectricblue!30}

\newcommand{\hldb}[1]{%
    {%
    \sethlcolor{lightdelectricblue}%
    \hl{#1}%
    }%
}

\definecolor{RowHighlight}{gray}{0.9}
\definecolor{darkblue}{rgb}{0,0.08,0.45}
\definecolor{Gray}{gray}{0.9}
\hypersetup{citecolor=darkblue, linkcolor=darkblue}

\newcommand{\txt}{\mathtt{inst}}
\newcommand{\img}{\mathtt{img}}

\newcommand{\cmark}{\ding{51}}%
\newcommand{\xmark}{\ding{55}}%

\newcommand{\llname}{ReGUIDE: Data Efficient GUI Grounding via \\Spatial Reasoning and Search}
\newcommand{\lname}{Data Efficient GUI Grounding via Spatial Reasoning and Search\xspace}

\newcommand{\mmname}{Reasoning Graphical User Interface Grounding for Data Efficiency\xspace}
\newcommand{\sname}{ReGUIDE\xspace}
\newcommand{\placeholder}[1]{{\color{lightgray}\lipsum[1]}}

\title{\llname}

\author{%
  Hyunseok Lee$^{1,2}$\thanks{Work done during an internship at NAVER Cloud},~~Jeonghoon Kim$^{1,2}$,~~Beomjun Kim$^{1}$,~~Jihoon Tack$^{1}$,~~Chansong Jo$^2$\\
  \textbf{Jaehong Lee$^2$,~~Cheonbok Park$^{1,2}$,~~Sookyo In$^2$,~~Jinwoo Shin$^{1,}$\thanks{Equal correspondence.}~,~~Kang Min Yoo$^{2,\dagger}$} \\
 KAIST$^1$, NAVER Cloud$^2$\\
  \texttt{\{hs.lee, jinwoos\}@kaist.ac.kr}, \texttt{kangmin.yoo@navercorp.com} \\  
}

\begin{document}

\maketitle

\input{sections/0.abstract}
\input{sections/1.introduction}
\input{sections/2.relatedwork}

\input{sections/3.method}

\input{sections/4.experiment}

\input{sections/5.conclusion}

\medskip

\bibliographystyle{abbrvnat}
\bibliography{ref}



\newpage

\input{sections/6.appendix}

\end{document}

%% file: sections/0.abstract.tex
\begin{abstract}

Recent advances in Multimodal Large Language Models (MLLMs) have enabled autonomous agents to interact with computers via Graphical User Interfaces (GUIs), where accurately localizing the coordinates of interface elements (e.g., buttons) is often required for fine-grained actions.
However, this remains significantly challenging, leading prior works to rely on large-scale web datasets 
to improve the grounding accuracy.
In this work, we propose \mmname (\sname), a novel and effective framework for web grounding that 
enables MLLMs to learn data efficiently
through self-generated reasoning and spatial-aware criticism. 
More specifically, \sname learns to (i) self-generate a language reasoning process for the localization via online reinforcement learning, and (ii) criticize the prediction using spatial priors that enforce equivariance under input transformations.
At inference time, \sname further boosts performance through a test-time scaling strategy, which combines spatial search with coordinate aggregation.
Our experiments demonstrate that \sname significantly advances web grounding performance across multiple benchmarks, outperforming baselines with substantially fewer training data points (that is, only 0.2\% samples compared to the best open-source baselines).

\end{abstract}

%% file: sections/1.introduction.tex
\section{Introduction}

Graphical User Interface (GUI) agents—i.e., Multimodal Large Language Model (MLLM) agents that interpret visual screen contents and generate web actions in natural language to navigate the web environment—have shown promising capabilities in web navigation tasks as MLLMs continue to improve in general decision-making ability \citep{openai2024gpt4ocard,anthropic2024computeruse, qin2025uitars, zheng2024seeact}. Despite the recent significant efforts, however, a substantial gap remains between the performance of these agents and that of proficient human users \citep{chen2025toward}, particularly in complex or long-horizon tasks \citep{zhang2025breaking}. This gap arises mainly from two challenges: limited visual understanding of web pages \citep{wang2024guisurv} and insufficient web domain-specific decision-making ability \citep{gou2024uground}. To address this, prior work typically decomposes the problem into two parts: (i) building an MLLM that can interpret the visual input \citep{zheng2024seeact} and (ii) employing a language model that performs decision-making and planning based on the MLLM's interpretation \citep{qin2025uitars, hong2024cogagent}, where the major bottleneck lies in the visual understanding component \citep{gou2024uground}.

Building an MLLM that can predict the exact pixel coordinates of target region on the screen (e.g., buttons) has shown great promise in providing the visual understanding required by another LLM for effective decision-making \citep{gou2024uground}. However, grounding remains challenging as it requires fine-grained skills such as reasoning and the manifestation of spatial understanding through token-based predictions \citep{ma2025deepperception,zhao2025robustness}. Recent approaches have emphasized the importance of large-scale, high-quality image–instruction pairs for training MLLMs through supervised fine-tuning (SFT) \citep{lin2024showui, gou2024uground}. \citet{gou2024uground} proposes synthesizing diverse referring expressions that can express the same object in several views. However, such SFT models depend on costly data curation to perform well, which poses a major scalability challenge \citep{huang2025visionr1}.

To this end, we focus on extracting rich information from web image data to achieve data-efficient grounding by learning to explain the coordinate prediction process in natural language and by leveraging spatial priors to maximize the utility of visual input.

\textbf{Contribution.} 
We propose \mmname (\sname), a novel and effective GUI coordinate grounding method for web agents.
Specifically, \sname is composed of a two-stage training: (i) self-generation of image explanation through its own reasoning, and (ii) criticism of the localization prediction via spatial priors. 
First, \sname learns to generate the language description of the given web image that can guide itself to correctly predict the coordinate, where this prediction accuracy is used as the reward for online RL.
Then, by leveraging a spatial prior—i.e., the fact that augmentations such as cropping lead to equivariant changes in the target coordinates—the model criticizes its predictions by ensuring consistent outputs under the same language description across augmented image–coordinate pairs.

Furthermore, we introduce an inference-time scaling strategy for \sname, which integrates spatial search with an LLM test-time scaling scheme.
Specifically, the model generates multiple localization predictions and crops the region where the target is likely to exist based on the predictions. 
In this region, we generate multiple coordinate candidates, 
then aggregate the candidates into a single coordinate 
via statistical voting strategy (i.e., Kernel Density Estimation~\citep{rosenblat1956remarks} (KDE)-based voting).

We demonstrate the effectiveness of \sname through evaluations on multiple web-grounding datasets and agent-setting benchmarks. Notably, \sname enhances web coordinate grounding performance beyond prior methods, achieving the state-of-the-art performance in web grounding. For instance, \sname improves the grounding accuracy of Qwen-2.5-VL-3B from 55.5\% to 87.3\% on \textsc{Screenspot~\citep{cheng2024screenspot}} and from 23.9\% to 44.3\% on \textsc{Screenspot-Pro~\citep{li2025screenspotpro}}. Furthermore, our experimental results show that \sname indeed guides the GUI agent to improve the overall decision-making ability, showing a significant performance improvement in agentic tasks.  

%% file: sections/2.relatedwork.tex
\section{Related Work}
\label{sec:related_work}

\textbf{Graphic User Interface (GUI) grounding.} 
Recent advances in pixel-level GUI grounding have demonstrated mapping natural language instructions to screen coordinates without relying on HTML or DOM structures~\citep{shaw2023pixels}. Several prior works~\citep{gou2024uground,xu2024aguvis,wu2024atlas, cheng2024seeclick, lin2024showui} train on over a million synthesized screenshots and achieve superior performance. ~\citet{gou2024uground} further shows that synthesizing and augmenting text instructions for each image element can lead to further improvements. However, current approaches depend on massive annotated datasets, incurring substantial computation and labeling costs, and neither leverages the reasoning capabilities of large language models to improve localization under data-scarce or out-of-distribution conditions.

\textbf{Reinforcement learning in MLLM.}
Reinforcement learning has emerged as a powerful mechanism to fine‐tune multimodal large language models via self‐improvement and feedback. Self‐Critical Sequence Training~\citep{rennie2017selfcriticim} pioneered policy‐gradient optimization for image captioning. Building on this paradigm, RLHF‐V~\citep{yu2024rlhfv} formulates multimodal RLHF under constrained optimization to jointly maximize helpfulness and minimize unsafe outputs. More recently, Vision‐R1 \citep{huang2025visionr1} generates synthetic chain‐of‐thought trajectories and applies iterative policy optimization to boost multimodal math reasoning. While UI-R1~\citep{lu2025uir1} and GUI-R1~\citep{xia2025guir1} are concurrent works that solves the GUI grounding problem with reinforcement learning, \sname further explores the grounding by utilizing the spatial prior and test-time spatial search. As a result, \sname significantly outperforms previous models in grounding evaluations as shown in Appendix.

\textbf{Test-time scaling.} 
Recent works explored that scaling test-time computation, such as best-of-N sampling, can be even better than scaling train-time computation for performance \citep{snell2024scaling}. Specifically, test-time scaling strategies improve LLM performance by generating numerous candidate outputs and selecting the best \citep{snell2024scaling,lee2025revise,hosseini2024vst}. To enhance decision-making, external verifiers are often employed to evaluate and refine these outputs \citep{hosseini2024vst}. In the localization task, recent work framed the localization as a search problem and suggested a test-time scalable strategy \citep{wu2024vstar}.

%% file: sections/3.method.tex
\input{resource/fig_method}
\section{\lname}
\label{sec:method}

In this section, we present \mmname (\sname), a Graphical User Interface (GUI) grounding (i.e., coordinate prediction) training framework that self-generates the language description of the image through reasoning, and criticizes the prediction with spatial priors.
We present the core training method in Section \ref{sec:train}, and then the test-time search method in Section \ref{sec:test-time}.
The overview of \sname is depicted in Figure \ref{figure:method}.

\textbf{Problem setup.}
We describe the problem setup of our interest, namely the GUI grounding: training an MLLM that can predict the coordinate of the interface (e.g., `close button') in the web image based on the given instruction (e.g., `close the window').
Formally, given an input instruction $x_\txt$ and image $x_\img$, the MLLM $\mathcal{M}$ generates the output which consists of reasoning path $r$ and predicted coordinate $\mathbf{c}=(c_1,c_2)$, i.e., $r,\mathbf{c} \sim \mathcal{M}(\cdot|x_\txt,x_\img)$. 
Here, we denote the ground-truth region in the pixel coordinate space corresponding to $(x_\txt, x_\img)$ as $\mathcal{X}_\mathtt{gt} \subseteq [0, W] \times [0, H]$, where $W$ and $H$ are the width and height of $x_\img$.
Any coordinate $\mathbf{c} \in \mathcal{X}_\mathtt{gt}$ is considered a correct prediction, as all such points result in the same GUI behavior (e.g., triggering the intended button or link).

\subsection{\sname: \mmname}
\label{sec:train}

We describe the core training pipeline of \sname. The key idea is to extract rich information from web images, enabling the model to generalize even with a relatively small training dataset. To this end, we train the model to self-explain the reasoning behind its predictions, as learning an explicit reasoning process—rather than merely memorizing input-output patterns—is crucial for generalization \citep{nye2021show, lampinen2022can}. Building on this learned reasoning, we further refine the predictions through a criticism step, which assesses the consistency of the output under image augmentations while maintaining the same reasoning trace.

\textbf{Learning to explain GUI images via reasoning.} At the first stage, we train the LLM to reason about the exact coordinate of the instruction on the screen, where we employ online reinforcement learning (RL) to make MLLM self-evolve by using grounding accuracy as a reward. This enables the model to generate its own text reasoning and evolve without relying on externally provided language descriptions, thus saving the language annotation cost.
RL also tends to generalize better than supervised fine-tuning (SFT), which may suffer from memorization \citep{nye2021show}. 

For a given instruction $x_\txt$ and an image $x_\img$, we define two reward functions to train MLLM with RL. 
Specifically, we consider (i) the result of ground truth prediction as the primary reward and (ii) an auxiliary formatting reward that encourages the model to wrap its reasoning trace between `\texttt{<think>}' and `\texttt{</think>}' tags, ensuring that the generated reasoning follows the desired format for downstream usage by following recent work in LLM reasoning \citep{huang2025visionr1, guo2025deepseekr1}. 
Formally, we define the reward function $R(\cdot,\cdot)$ as follows:
\begin{equation}
    R(r, \mathbf{c}) = \mathbf{1}[\mathbf{c} \in \mathcal{X}_\mathtt{gt}]
    + \lambda \cdot \mathbf{1}[\text{format}(r) = \texttt{<think>} \cdots \texttt{</think>}],
\end{equation}
where $\mathbf{1}[\cdot]$ denotes the indicator function that returns one if the condition holds and zero otherwise, $\text{format}(r)$ extracts the outermost structure of the generated reasoning trace to check whether it is properly enclosed between \texttt{<think>} and \texttt{</think>}, and $\lambda\in\mathbb{R}^{+}$ is a hyperparameter that balances the contribution of the formatting reward, where we use $\lambda=0.1$ throughout the experiment.

The overall RL objective is to maximize the expected reward:
\begin{equation}
    \max_{\mathcal{M}} \; \mathbb{E}_{x_\txt, x_\img} \, \mathbb{E}_{(r, \mathbf{c}) \sim \mathcal{M}(\cdot \mid x_\txt, x_\img)} \left[ R(r, \mathbf{c}) \right].
\end{equation}
To this end, we adopt Group Relative Policy Optimization (GRPO) \cite{shao2024grpo} as our RL algorithm, as GRPO enhances both efficiency and stability by sampling multiple candidate outputs per input and computing group-relative rewards.
This overall process allows the model to explore diverse reasoning strategies and progressively reinforce those that contribute to accurate grounding, without relying on explicit reasoning supervision.

\textbf{Learning to predict consistent coordinates under transformations.} While the RL stage enables the model to predict coordinates within the correct ground truth region, it does not necessarily guide the model toward predicting the most precise or stable point, typically the center of the target element (as the target UI element is annotated with a bounding box, making the center a natural target). Additionally, the model may be sensitive to variations in scale and cropping, which can degrade performance in real-world scenarios where UI elements may appear at different sizes or positions.

To address these limitations, we introduce a further training phase that enforces spatial precision and view consistency. Specifically, we continually train the MLLM to predict the center of the ground-truth region, while jointly enforcing consistent predictions under image transformations through spatial priors. In particular, we focus on constraining consistency between the model’s predictions on the global view (full image) and the local view (a zoomed-in crop covering the ground-truth region). Notably, such global-local view consistency has been identified as a key factor in the success of prior self-supervised vision methods \citep{caron2021dino, oquab2023dinov2, hjelm2018learning}.
Furthermore, we make the local view follow the reasoning process of the global view (as it typically contains relative spatial cues), forming a self-distillation of the reasoning process.

Formally, we collect two data points, namely the global view data $\mathbf{d}_{\mathtt{global}}$ and local view data $\mathbf{d}_{\mathtt{local}}$, where each data point consists of image, instruction, reasoning, and target coordinates. 
Here, both data points consist of the same instruction and reasoning, but with different images and target coordinates (due to the spatial transformation). 
To ensure high-quality reasoning, we only use samples that correctly predict the target region in the global image $x_\img$, i.e., $r$ from $(r, \mathbf{c}) \sim \mathcal{M}(\cdot \mid x_\txt, x_\img)$ such that $\mathbf{c} \in \mathcal{X}_{\mathtt{gt}}$.
For the global view data, we use the non-transformed image $x_\img$ and the center of the ground truth region $\mathtt{center}(\mathcal{X}_{\mathtt{gt}})$, while for the local view data, we use a randomly cropped image $\mathtt{crop}(x_\img)$ and the center of the transformed ground truth region, denoted as $\mathtt{center}(\mathcal{X}_{\mathtt{gt}}^{\mathtt{crop}})$. Namely, the global and local view data are defined as:
\begin{equation*}
\mathbf{d}_{\mathtt{global}}:=\big(x_\txt, x_\img, r, \mathtt{center}(\mathcal{X}_{\mathtt{gt}})\big),\quad
\mathbf{d}_{\mathtt{local}}:=\big(x_\txt, \mathtt{crop}(x_\img), r, \mathtt{center}(\mathcal{X}_{\mathtt{gt}}^{\mathtt{crop}})\big).
\end{equation*}
For training, the model processes the original global view and the newly created local view in the same batch, and is supervised via next-token prediction loss to generate the updated coordinates as text tokens:
\begin{equation}
    \min_\mathcal{M} ~\mathcal{L}(\mathbf{d}_{\text{global}}) + \mathcal{L}(\mathbf{d}_{\text{local}}), \quad \text{where} \quad \mathcal{L}(\mathbf{d}) = - \log \mathcal{M}(r, \mathbf{c} \mid x_\txt, x_\img'),
\end{equation}
where $\mathbf{d} = (x_\txt, x_\img', r, \mathbf{c})$ and $x_\img'$ denotes the input image used in the given view (i.e., $x_\img$ for the global view and $\mathtt{crop}(x_\img)$ for the local view).

\subsection{Test-time Scaling with Spatial Search and Kernel Density Estimation Based Aggregation}
\label{sec:test-time}

Despite their strong reasoning capabilities via natural text, LLMs inherently struggle with coordinate prediction due to their limited awareness of ordinal relationships among numeric tokens. To alleviate the issue, we propose a scalable inference method for models trained with \sname. Specifically, we introduce a two-stage inference method, namely, composed of the (i) cropping stage and the (ii) voting stage. Here, the key idea for each stage is to (i) zoom in on the image where the UI element is likely to exist, and (ii) predict multiple coordinates on the zoomed image, which are aggregated into a single coordinate for a robust and confident prediction. An illustration of our test-time spatial search strategy is provided in Appendix.

\textbf{Cropping: Zooming into the UI element area.} To crop the region where the UI element is likely to exist, we first predict multiple coordinates on the full image. Then, we treat these predictions as samples from a probability distribution indicating the target's likely location. Here, we use Kernel Density Estimation (KDE) to analyze these initial samples and identify the region of highest prediction density—the area where the model is most consistently expected to be the target. Thus, we choose the highest density point as the center of the cropped Region of Interest (RoI), effectively allowing the model to "zoom in" on the most promising area. 

Concretely, given an input image $x_\img$ and instruction $x_\txt$, we sample $N$ initial predictions: $\mathcal{C}:=\{\mathbf{c}^{(i)}\}_{i=1}^N$, where $(r^{(i)}, \mathbf{c}^{(i)}) \sim \mathcal{M}(\cdot \mid x_\txt, x_\img)$.
Then, we apply KDE to these $N$ predictions to find the `center' $\mathbf{c}_{\mathtt{KDE}}$ of the predictions by summing 2D Gaussian kernels (with a pre-defined variance $\Sigma$) centered at each prediction $\mathbf{c}^{(j)}$ to estimate a density $S(z;\mathcal{C})$: 
\begin{equation}
    \mathbf{c}_{\mathtt{KDE}} = \operatorname*{argmax}_{z\in[0,W]\times[0,H]} S(z;\mathcal{C})
    \quad\text{where}\quad
    S(z;\mathcal{C})=\mathbb{E}_{\mathbf{c}\in\mathcal{C}}\Big[\exp\bigl(-\tfrac12(z-\mathbf{c})^{\!\top}\Sigma^{-1}(z-\mathbf{c})\bigr)\Big].
    \label{eq:kde}
\end{equation}
A fixed-size bounding box $\mathcal{X}_{\mathtt{RoI}}$ (with dimensions $W_{\mathtt{RoI}} \times H_{\mathtt{RoI}}$), centered at $\mathbf{c}_{\mathtt{KDE}}$, defines the region of interest. We crop the image deterministically to this region:
$\mathtt{RoI}(x_{\mathtt{img}}) = \mathtt{crop}(x_{\mathtt{img}};\, \mathbf{c}_{\mathtt{KDE}},\, W_{\mathtt{RoI}},\, H_{\mathtt{RoI}})$.

\textbf{Voting: Aggregating multiple votes within RoI.}
In the voting stage, we further refine the coordinate prediction within this RoI. Given the local zoomed-in view, these predictions are expected to be more precise -- the answer space is narrowed. We then reapply KDE to these new predictions within the RoI. This second application of KDE acts as a robust voting mechanism, aggregating the multiple refined predictions to determine the single coordinate with the highest probable point (i.e., the peak of the density). This two-stage searching strategy not only allows ReGUIDE to more precisely predict the final answer but also offers scalability: investing more computational resources in generating and evaluating samples for refinement generally leads to a more accurate result.

Formally, the model re-predicts $M$ new coordinates: 
$\mathcal{C}_{\mathtt{RoI}}:=\{\mathbf{c}_{\mathtt{RoI}}^{(i)}\}_{i=1}^N$, where $(r^{(i)}, \mathbf{c}_{\mathtt{RoI}}^{(i)}) \sim \mathcal{M}(\cdot \mid x_\txt, \mathtt{RoI}(x_{\mathtt{img}}))$.
Then, we use the same process as in \autoref{eq:kde} to predict the most confident center point $\mathbf{c}_{\mathtt{final}}$ within $\mathcal{C}_{\mathtt{RoI}}\cup\mathcal{C}$ by robustly aggregating coordinates with KDE:
\begin{equation}
\mathbf{c}_{\mathtt{final}}
=\operatorname*{argmax}_{z\in\mathcal{\mathcal{X}}} S(z;\mathcal{C}_{\mathtt{RoI}}\cup\mathcal{C}).    
\end{equation}

%% file: resource/fig_method.tex
\begin{figure*}[t]
\begin{subfigure}{0.5055\textwidth}
    \includegraphics[width=\textwidth]{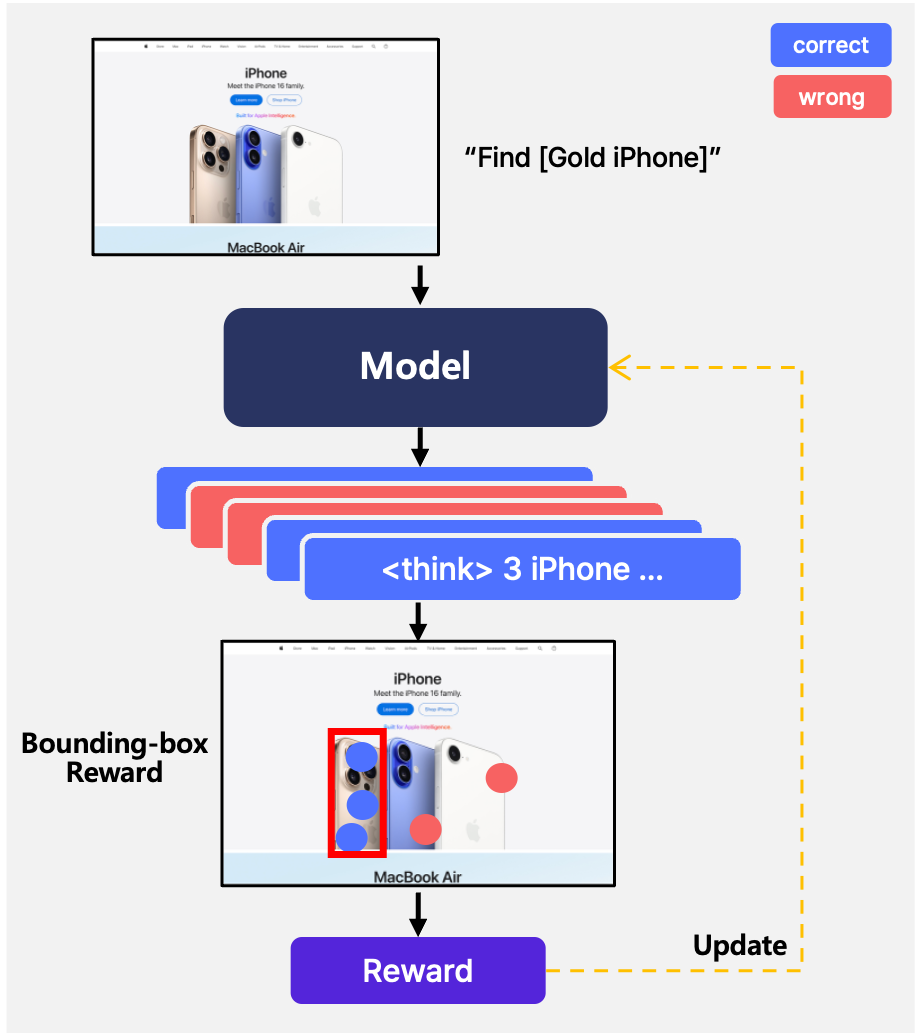}
    \caption{Learning to explain web images via reasoning}
    \label{figure:concept1}
\end{subfigure}
\begin{subfigure}{0.48\textwidth}
    \includegraphics[width=\textwidth]{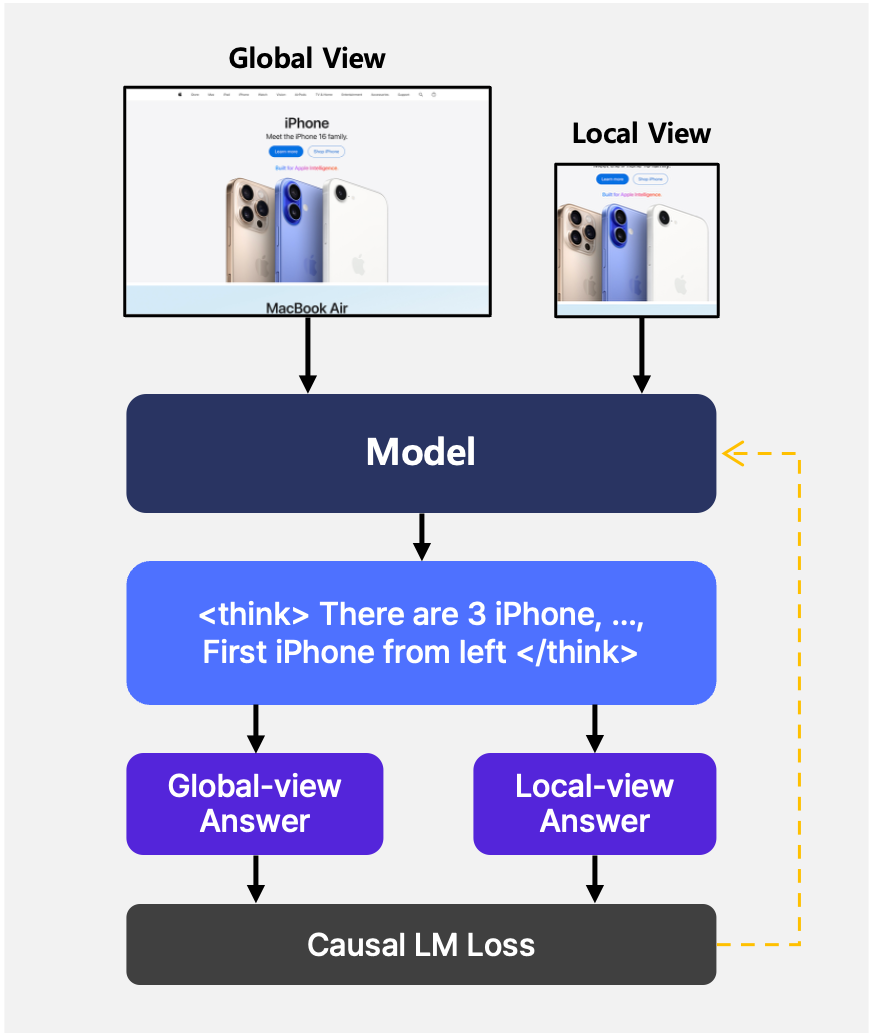}
     \caption{Learning to predict consistent coordinates}
    \label{figure:concept2}
\end{subfigure}
\caption{\textbf{Overview of training \sname.} \textbf{Left}: The model rolls out multiple (reasoning, coordinate) pairs; rewards the point that falls inside the ground-truth box. \textbf{Right}: Trains the model on paired full-image and cropped views, sharing the reasoning tokens while adjusting the coordinate, which provides multi-view consistency and improves grounding performance.}
\label{figure:method}
\end{figure*}

%% file: sections/4.experiment.tex
\section{Experiments}
\label{sec:experiments}
We provide an empirical evaluation of \sname by investigating the following questions:

\begin{itemize}[leftmargin=*,topsep=0.0pt,itemsep=.5pt]
\item Can \sname enhance GUI grounding performance? (Table~\ref{tab:main} and Table~\ref{tab:domain})
\item Can \sname improve overall GUI agent performance? (Table~\ref{tab:agent})
\item How does the MLLM perform reasoning based on the given instruction and image? (Table \ref{tab:gen_example})
\item Do the proposed components enhance the grounding performance? (Table \ref{tab:training_ablation} and Table~\ref{tab:scaling})
\end{itemize}

Before answering each question, we outline the experimental protocol (more details in Appendix \ref{app:experiment_detail}).

\textbf{Evaluation setup.} In the main results, we mainly report the grounding accuracy (\%) as a metric. The prediction is counted as correct when the predicted point lies inside the ground-truth bounding box. During evaluation, we generate a prediction via greedy decoding. For the searching strategy of \sname we used prediction sampling $N=M=16$ samples with temperature $T=1.0$, crop bounding box size as $W_{\mathtt{RoI}}=H_\mathtt{RoI}=840$, and KDE variance $\Sigma=0.01$. We evaluate \sname and baselines on \textsc{ScreenSpot} \citep{cheng2024screenspot}, \textsc{ScreenSpot-v2} \citep{wu2024atlas}, \textsc{ScreenSpot-Pro} \citep{li2025screenspotpro}, evaluation benchmarks for GUI grounding. Especially, \textsc{ScreenSpot-Pro} is a more challenging construct with a high-resolution image (i.e., up to 3840 $\times$ 2160). The agentic evaluation setting is in Section~\ref{sec:agentic}. 

\textbf{Training setup.} For the main experiment, we train \sname on Qwen-2.5-VL 3B/7B. To demonstrate \sname's data efficiency, we utilize only a 20k subset of UGround~\citep{gou2024uground} dataset constituting approximately 0.2\% of its full set. Learning to explain web images via reasoning, we use the GRPO \citep{shao2024grpo} algorithm with two rewards, accuracy reward and output-format compliance reward, and $\lambda=0.1$ for formatting reward weight. For learning to predict consistent coordinates under transformations, where the local view is a random crop of up to 30\% of the original area while preserving aspect ratio. More setups in Appendix~\ref{appendix:training_setup}.

\textbf{Baselines.} We compare our method against several baselines, which fall into two categories. First, we consider the model Qwen-VL-2.5 3B/7B supervised fine-tuned (SFT) on the identical dataset with \sname (i.e., 20k UGround subset). SFT optimises a coordinate-only loss and does not generate reasoning. Additionally, we consider existing proprietary MLLM and GUI grounding models: GPT-4o~\citep{openai2024gpt4ocard}, Claude3.7~\citep{anthropic2024claude}, SeeClick~\citep{cheng2024seeclick}, OS-Atlas-7B~\citep{wu2024atlas}, AGUVIS-7B~\citep{xu2024aguvis}, and UGround~\citep{gou2024uground}.

\input{resource/tab_main}

\subsection{Main Results}

We first present the main result by comparing the GUI grounding performance with other baselines. Here, we mainly compare \sname with finetuning baselines with the same model and dataset, and other open and closed models. 

As shown in Table~\ref{tab:main}, we present the GUI grounding performance of \sname compared to other baselines. We first compare models trained with the same base architecture (Qwen-2.5-VL) and dataset (a 20k subset of UGround). Our proposed method, \sname, significantly outperforms supervised fine-tuning (SFT) baselines across all evaluated scenarios. For instance, \sname achieves an accuracy of 87.3\% on \textsc{ScreenSpot} with the 3B model. Similar improvements are observed for the more challenging \textsc{ScreenSpot-Pro} benchmark, where \sname boosts performance from 23.9\% to 44.3\%. Also in \textsc{ScreenSpot-v2} and dataset, \sname outperforms baseline and SFT model with a meaningful margin.

\textbf{Comparison with open-sourced and proprietary models.}
Beyond models trained with the same data and architecture, we compare \sname to state-of-the-art open-sourced and proprietary models, including SeeClick~\citep{cheng2024seeclick}, Os-Atlas-7B~\citep{wu2024atlas}, AGUVIS-7B~\citep{xu2024aguvis}, UGround~\citep{gou2024uground}, GPT-4o~\citep{openai2024gpt4ocard}, and Claude~\citep{anthropic2024claude}.  
\sname, despite being trained on only a 20k subset of UGround, achieves competitive or superior performance.  
Specifically on \textsc{ScreenSpot}, our 3B model surpasses UGround-7B (87.3\% > 86.3\%) and \sname's 7B model's superior results against other baselines. Also, \sname consistently outperforms other baselines in \textsc{ScreenSpot-v2}, too.
\sname shows the most remarkable performance on \textsc{ScreenSpot-Pro}. \sname performs accuracy of 45.1\%, significantly outperforming UGround-7B (31.1\%).

\textbf{Training Efficiency of \sname.}
These results demonstrate that \sname not only surpasses supervised finetuned within the same training dataset but also establishes a new competitive standard against much larger, heavily trained open and closed models.  
Importantly, \sname achieves these improvements \emph{using only a small fraction of the data (a 20k subset of UGround)} compared to other models (i.e., UGround: 10M, AGUVIS: 1M), highlighting its data efficiency and scalability.
By combining reinforcement learning, transform consistency learning, and scalable searching inference, \sname achieves a strong performance.

\subsection{Additional Analysis and Ablation}
\input{resource/tab_domain}
\textbf{Generalizability over several domains.} 
We further analyze the robustness and generalizability of \sname across various GUI domains in multiple device environments (Mobile, Desktop, Web) and UI element types (Text, Icon). While our main results in Table~\ref{tab:main} report the average accuracy on the \textsc{ScreenSpot}, here we provide a more fine-grained domain-specific evaluation. As shown in Table~\ref{tab:domain}, \sname consistently achieves superior or competitive performance compared to the state-of-the-art open-sourced base model (i.e., Uground~\citep{gou2024uground}). These results indicate that the training schemes of \sname effectively enhance its ability to generalize across diverse environments. Notably, \sname particularly shows strong performance for Icon elements within the Mobile and Desktop domains. We believe that the multi-view consistency training attributes the results, which likely enhances the model's understanding of the local visual features present in iconic representations.

\input{resource/tab_gen_example}

\textbf{Relative spatial cue of reasoning generations.} 
We analyzed the generated reasoning path of \sname and the associated spatial predictions. First, we observe that the model naturally decomposes GUI localization tasks into interpretable reasoning steps, referencing relative spatial information. As shown in Table~\ref{tab:gen_example}, the model explicitly identifies intermediate visual landmarks (e.g., a labeled section of "Top free apps") before specifying precise coordinates based on relative positions (e.g., "first row, first column"). This structured, relative reasoning pattern supports the models' ability to generalize across varying UI layouts.

\input{resource/fig_confidence}
\textbf{Natural Gaussian distribution of ground likelihood.} 
As an additional qualitative analysis, we investigate the likelihood distributions of coordinates surrounding the initial predicted coordinates. By calculating the likelihood of coordinates with same generated reasoning.
As shown in Figure~\ref{figure:confidence}, the likelihoods of the surrounding coordinates are a surprisingly smooth, Gaussian-shaped distribution. 
This Gaussian pattern indicates that the MLLM implicitly captures continuous spatial uncertainty, despite operating entirely in token space.
This observation provides a strong empirical justification for our proposed Gaussian-weighted inference strategy, that naturally emerging confidence distributions to enhance coordinate prediction accuracy and robustness.

\input{resource/tab_training_ablation}
\textbf{Analysis of training components.} We perform an analysis on each component of \sname, specifically spatial reasoning (Reasoning), learn consistency under transformation (Consistency), and test-time searching strategy (Searching). As shown in Table~\ref{tab:training_ablation}, each component plays an important role, leading to gradual and significant improvements when applied sequentially. Additional comparisons of alternative RL policy-optimization algorithms are provided in Appendix \ref{appendix:other_rl}.

\input{resource/tab_scaling}
\textbf{Analysis of inference components.} We perform ablation to prove the effectiveness of each component in the two-stage inference time searching strategy. As shown in Table~\ref{tab:scaling}, both the crop stage and the voting stage contribute to improvement in performance. It is notable that in \textsc{Screenspot-Pro}, the improvement induced by crop stage is huge, which indicates that localization plays a crucial role for the proper understanding of high-resolution images.

\input{resource/fig_ablation_searching}
\textbf{Ablation on inference time searching strategy.} We perform ablation for three hyperparameters on inference time scaling strategy, that is, crop size ($W_\mathtt{RoI}$), generation samples ($N$), and temperature ($T$). As shown in Figure~\ref{fig:ablation_all}, increasing $N$ gradually improves the grounding performance, demonstrating that our strategy is scalable. Unless stated otherwise, we adopt the default hyperparameters $N=16$, $L=840$, and $T=1.0$, which balance performance and cost.

\input{resource/tab_agent}
\subsection{Agentic Settings Experiments} \label{sec:agentic}
To verify whether better grounding transfers to real tasks, we plug each grounding model with \sname and use a high-level planner (GPT-4o). Here, we report task-success rate (\%) on five offline suites: Multimodal-Mind2Web-cross domain, -cross task,  and -cross website~\citep{deng2023mindweb}, AndroidControl-Low and -High~\citep{li2024androidcontrol}. Each benchmark has a different domain of agent settings (i.e., desktop web navigation and mobile-app control). For every episode, the planner emits a natural-language action plan; the grounding module converts each step into pixel coordinates, which the environment executes. Success is recorded when the episode reaches the goal state. All prompts, settings, and frameworks are followed by UGround~\citep{gou2024uground}. The detailed settings are described in Appendix~\ref{app:experiment_detail}.

As shown in Table~\ref{tab:agent}, improved grounding performance made by \sname leads to better agent performance. With only 3B parameters, \sname matches or edges out the larger UGround-7B on Multimodal-Mind2Web and Android Control-High. The 7B variant of \sname consistently outperforms other baselines (i.e., +1.3\% in Multimodal-Mind2Web cross task). These gains, achieved with identical planners and action budgets, substantiate that the grounding improvements provided by self-evolutionary RL and global–local consistency directly enhance full-task completion.

%% file: resource/tab_main.tex
\begin{table*}[t] 
\caption{Accuracy (\%) for \sname (Ours) and other baselines, including GPT-4o \citep{openai2024gpt4ocard}, Claude 3.7 \citep{anthropic2024claude}, SeeClick~\citep{cheng2024seeclick}, OS-Atlas-7B~\citep{wu2024atlas}, AGUVIS-7B~\citep{xu2024aguvis}, UGround \citep{gou2024uground}, Qwen-2.5-VL \citep{bai2025qwen25vl}, and \sname (Ours). We evaluate on three web-grounding benchmarks: \textsc{ScreenSpot}~\citep{cheng2024screenspot}, \textsc{ScreenSpot-v2}~\citep{wu2024atlas}, and \textsc{ScreenSpot-Pro}~\citep{li2025screenspotpro}. The \textit{Proprietary Models} include proprietary systems, while \textit{Open-Sourced} cover public research models. Models supervised-finetuned (SFT) with the same architecture (i.e., Qwen-2.5-VL) and dataset (i.e., a 20K subset of UGround) are grouped separately for controlled comparisons. Bold indicates the best result within each group.}
\label{tab:main}
\small
\centering
\resizebox{\textwidth}{!}{
\begin{tabular}{lcccc}
\toprule
Methods & Data Size & \textsc{ScreenSpot}~\citep{cheng2024screenspot} & \textsc{ScreenSpot-v2}~\citep{wu2024atlas}& \textsc{ScreenSpot-Pro}~\citep{li2025screenspotpro} \\
\midrule
 \multicolumn{5}{l}{\textit{Proprietary Models}}\vspace{0.02in} \\
\phantom{00}GPT-4o \citep{openai2024gpt4ocard} & - & 18.3 & 16.6 & \phantom{0}0.8 \\
\phantom{00}Claude 3.7 \citep{anthropic2024claude} & - & 82.1 & 87.6  & 27.7 \\
\midrule 
\multicolumn{5}{l}{\textit{Open-Sourced}}\vspace{0.02in} \\
\phantom{00}SeeClick \citep{cheng2024seeclick} & \phantom{0}1M & 53.4 & 55.1 & \phantom{0}1.1 \\
\phantom{00}OS-Atlas-7B \citep{wu2024atlas}& 13M &  82.5 & 84.1 & 18.9 \\
\phantom{00}AGUVIS-7B \citep{xu2024aguvis} & \phantom{0}1M & 84.4 & 85.8 & 22.9 \\
\phantom{00}UGround-2B \citep{gou2024uground}& 10M & 77.7 & 81.4 & 26.6 \\
\phantom{00}UGround-7B \citep{gou2024uground}& 10M & 86.3 & 89.1 & 31.1 \\
\midrule
\multicolumn{5}{l}{\textit{Trained with same model and dataset}}\vspace{0.02in} \\
\phantom{00}Qwen-2.5-VL-3B \citep{bai2025qwen25vl}& - & 55.5 & 70.4& 23.9 \\
\phantom{0000}+ SFT & 20K & 56.8 & 56.5 & 11.6 \\
\rowcolor{RowHighlight}\phantom{0000}\textbf{+ \sname (Ours)} & 20K & \textbf{87.3} & \textbf{89.7}& \textbf{44.3} \\
\arrayrulecolor{black!40}\midrule
\phantom{00}Qwen-2.5-VL-7B \citep{bai2025qwen25vl} & - & 84.7 & 82.6 & 29.0 \\
\phantom{0000}+ SFT& 20K & 84.9 & 88.9 & 25.9 \\
\rowcolor{RowHighlight}\phantom{0000}\textbf{+ \sname (Ours)}& 20K & \textbf{90.1} & \textbf{91.5}& \textbf{44.4} \\
\arrayrulecolor{black}\bottomrule
\end{tabular}
}
\vspace{-.1in}
\end{table*}

%% file: resource/tab_domain.tex
\begin{table}[t]
\centering
\caption{Grounding Accuracy (\%) for \sname (Ours) and UGround baselines\citep{gou2024uground} on the \textsc{ScreenSpot}~\citep{gou2024uground} domain split. Results are broken down by device class (Mobile, Desktop, Web, and UI element type (Text, Icon). The right-most column reports the overall average. Bold indicates the best result within each column.}
\label{tab:domain}
\small
\begin{tabular}{lc cc cc cc c}
\toprule
Model& Data Size & \multicolumn{2}{c}{Mobile} & \multicolumn{2}{c}{Desktop} & \multicolumn{2}{c}{Web} & Average\\
\cmidrule(r){3-4} \cmidrule(r){5-6} \cmidrule(r){7-8}
 && Text & Icon & Text & Icon & Text & Icon \\
\midrule
UGround-2B~\citep{gou2024uground} &10M & 89.4 & 72.0 & 88.7 & 65.7 & 81.3 & 68.9 & 77.7\\
UGround-7B~\citep{gou2024uground} &10M & 93.0 & 79.9 & \textbf{93.8} & 76.4 & 90.9 & \textbf{84.0} & 86.3\\
\rowcolor{RowHighlight}\textbf{\sname-3B} &20K & 94.9 & 81.2 & \textbf{93.8} & 83.6 & 87.0 & 81.1 & 87.3\\
\rowcolor{RowHighlight}\textbf{\sname-7B} &20K & \textbf{95.2} & \textbf{89.5} & \textbf{93.8} & \textbf{83.6} & \textbf{91.3} & 83.5 & \textbf{90.1}\\
\bottomrule
\vspace{-0.25in}
\end{tabular}
\end{table}

%% file: resource/tab_gen_example.tex
\begin{table}[t]
  \centering
  \caption{Example of \sname{}’s generated reasoning and predicted coordinate.}
  \label{tab:gen_example}
  \begin{tabular}{c p{0.6\textwidth}}
    \toprule
    \raisebox{-0.87\height}{\includegraphics[width=0.33\textwidth]{images/img_wechat.jpg}}
    &
    \begin{minipage}[t]{\linewidth}
      \textbf{Target Prompt:} Choose WeChat.\\[0.4em]
      \textbf{Response:} \texttt{<think>} The task is to find the coordinate of the “WeChat” app icon in the image.
      The image shows a section labeled “Top free apps” with icons and names of apps.
      The ``WeChat'' app icon is located in the \hldb{first row, first column} of this section.
      \texttt{</think> <answer>(110,437)</answer>}
    \end{minipage}
    \\ \bottomrule
  \end{tabular}
  \vspace{-0.15in}
\end{table}

%% file: resource/fig_confidence.tex
\begin{wrapfigure}[13]{r}{0.36\columnwidth}
  \vspace{-0.18in}
  \includegraphics[width=0.35\columnwidth]{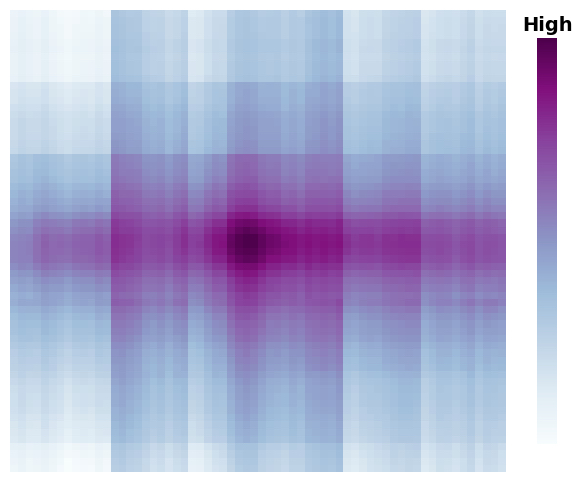}
  \caption{Likelihood of coordinate tokens spreaded gaussian naturally.}
  \label{figure:confidence}
\end{wrapfigure}

%% file: resource/tab_training_ablation.tex
\begin{table*}[t] 
\caption{Contribution of each proposed component of \sname on GUI grounding. We tested all three training components trained with Qwen-2.5-VL-3B: learn spatial reasoning (Reason), learn consistency under transformation (Consistency), and test-time spatial searching (Search). We report the GUI grounding accuracy (\%) on \textsc{ScreenSpot} and \textsc{ScreenSpot-Pro} benchmarks.} 
\label{tab:training_ablation}
\vspace{-0.07in}
\small
\centering
\begin{tabular}{c >{\centering\arraybackslash}m{1.5cm} >{\centering\arraybackslash}m{1.5cm} >{\centering\arraybackslash}m{1.5cm} cc}
\toprule
Model Size & Reason & Consistency & Search &\textsc{ScreenSpot} \citep{cheng2024screenspot} & \textsc{ScreenSpot-Pro} \citep{li2025screenspotpro} \\

\midrule
\multirow{4}{*}{3B}&\xmark & \xmark & \xmark & 55.5 & 23.9 \\
&\cmark & \xmark & \xmark & 83.3 & 27.2 \\
&\cmark& \cmark & \xmark & 84.9 & 27.9 \\
&\cmark& \cmark & \cmark & \textbf{87.3} & \textbf{44.3} \\
\midrule
\multirow{4}{*}{7B}&\xmark & \xmark & \xmark & 84.7 & 29.0 \\
&\cmark & \xmark & \xmark & 85.6 & 27.0 \\
&\cmark& \cmark & \xmark & 89.1 & 30.1 \\
&\cmark& \cmark & \cmark &\textbf{90.1} & \textbf{44.4} \\
\bottomrule
\end{tabular}
\vspace{-.15in}
\end{table*}

%% file: resource/tab_scaling.tex
\begin{table*}
\caption{
Contribution of each proposed component for test-time scaling, namely, the cropping and voting. We report the
grounding accuracy (\%) on \textsc{ScreenSpot} and \textsc{ScreenSpot-Pro} benchmarks with \sname-3B. The bold indicates the best results.
} 
\label{tab:scaling}
\vspace{-0.1in}
\small
\centering

\begin{tabular}{>{\centering\arraybackslash}m{1.5cm} >
{\centering\arraybackslash}m{1.5cm} ccc}
\toprule
Cropping & Voting & \textsc{ScreenSpot}~\citep{cheng2024screenspot} & \textsc{ScreenSpot-Pro}~\citep{li2025screenspotpro} \\
\midrule
\xmark & \xmark & 84.3 & 27.9 \\
\cmark & \xmark & 81.7  & 42.7  \\
\cmark & \cmark & \textbf{87.3} & \textbf{44.3}  \\
\bottomrule
\end{tabular}
\vspace{-.25in}
\end{table*}

%% file: resource/fig_ablation_searching.tex
\begin{figure}[t!]

  \centering
  \begin{subfigure}{0.32\textwidth}
    \centering
    \includegraphics[width=\textwidth]{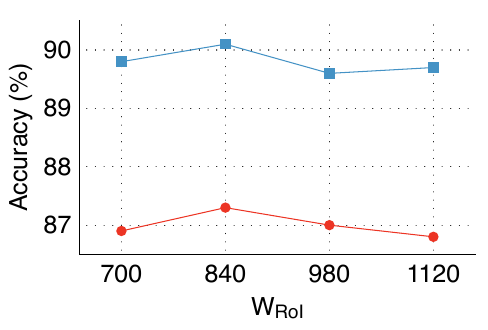}
    \caption{Effect of $W_{\mathtt{RoI}}$ (crop size).}
    \label{fig:ablation_L}
  \end{subfigure}
  \hfill
  \begin{subfigure}{0.32\textwidth}
    \centering
    \includegraphics[width=\textwidth]{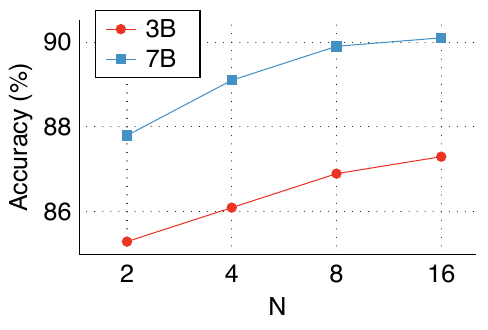}
    \caption{Effect of $N$ (num. samples).}
    \label{fig:ablation_N}
  \end{subfigure}
  \hfill
  \begin{subfigure}{0.32\textwidth}
    \centering
    \includegraphics[width=\textwidth]{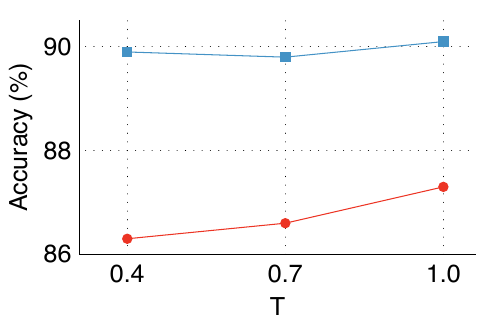}
    \caption{Effect of $T$ (temperature).}
    \label{fig:ablation_T}
  \end{subfigure}
  \caption{Ablation studies on grounding accuracy (\%) w.r.t.  
    (a) crop size $W_{\mathtt{RoI}}$,  
    (b) number of sampled predictions $N$, and  
    (c) decoding temperature $T$.}
  \label{fig:ablation_all}

\end{figure}

%% file: resource/tab_agent.tex
\begin{table*}[t] 
\caption{Task-success rate (\%) for \sname (Ours) and UGround~\citep{gou2024uground} baselines on the five offline-agent task benchmarks, i.e., Multimodal-Mind2Web (MM2W)~\citep{zheng2024seeact,deng2023mindweb}'s cross-domain, -task, -web, Android Control (AC)-low and -high ~\citep{li2024androidcontrol}. In every experiment, we used GPT-4o as a planner, while the listed models provide the grounding module.}
\label{tab:agent}
\small
\centering
\begin{tabular}{lc cccccc}
\toprule 
\multirow{2.5}{*}{Grounding Module}& \multirow{2.5}{*}{Data Size} & \multicolumn{3}{c}{MM2W~\citep{zheng2024seeact,deng2023mindweb}}& \multicolumn{2}{c}{AC~\citep{li2024androidcontrol}} \\
\cmidrule(lr){3-5} \cmidrule(lr){6-7}
 & & domain & task & web & high & low \\
\midrule
UGround-2B \citep{gou2024uground}& 10M & 47.7 & 48.6 & 47.6 & 65.0 & \textbf{50.0}\\
UGround-7B \citep{gou2024uground}& 10M & 48.5 & 50.7 & 48.1 & 66.2 & 49.8\\
\midrule
\rowcolor{RowHighlight}\textbf{\sname-3B (Ours)}& 20K & 48.8 & 50.5 & 47.7 & 66.2 & 49.8\\
\rowcolor{RowHighlight}\textbf{\sname-7B (Ours)}& 20K & \textbf{49.5} & \textbf{52.0} & \textbf{48.7} & \textbf{67.4} & \textbf{50.0}\\
\bottomrule
\end{tabular}
\vspace{-.07in}
\end{table*}

%% file: sections/5.conclusion.tex
\section{Discussion and Conclusion}
We proposed \sname, a novel and effective GUI grounding framework that significantly enhances the capabilities of Multimodal Large Language Models by enabling data-efficient learning through self-generated reasoning and spatial-aware criticism. Our key idea is leveraging online reinforcement learning for self-generating language reasoning and employing spatial priors to criticize predictions, and further boosting performance at inference time through a test-time scaling search strategy that integrates spatial search with coordinate aggregation. We demonstrated that \sname consistently outperforms other open-sourced baselines, even when trained with a tiny fraction of data, such as only 0.2\% of the samples used by the best open-sourced baselines. Crucially, these advances in data-efficient grounding translate to improved performance in downstream agentic tasks, highlighting \sname's potential to develop more capable and practical GUI agents.

\textbf{Future works and limitations.}\label{sec:future_limitation} We believe it will be an interesting future direction to train MLLM planners that can operate hierarchically with \sname, fully capitalizing on its precise grounding performance. Additionally, a potential limitation is that while \sname is a very effective method, it lacks an explicit safety framework to prevent malicious uses, such as hacking or spreading misinformation—a challenge common to many grounding models. However, future investigations could explore the research and development of robust safeguards.

%% file: sections/6.appendix.tex
\newpage
\appendix

\section{Experimental Details}
\label{app:experiment_detail}
In this section, we describe the experimental details of Section \ref{sec:experiments}, including \sname and baselines. 
\subsection{Dataset Details}
In this section, we describe the dataset we used in training and evaluation.
\begin{itemize}[itemsep=0.05in,leftmargin=0.2in,topsep=0.02in]
\item \textbf{UGround} Uground is a Graphical User Interface (GUI) grounding dataset that consists of 10M triplets of GUI image, element coordinates, and description of the element, which was synthesized by open MLLM, Llava-Next-13 B. Uground~\citep{gou2024uground} models are trained with their own synthesized web GUI dataset, which consists of 10 M image-instruction pairs. For training \sname, we randomly select a 20K image-instruction set from UGround~\citep{gou2024uground} dataset, which is approximately 0.2\% of the original dataset.

\item \textbf{\textsc{ScreenSpot}} ScreenSpot~\citep{cheng2024screenspot} is an evaluation benchmark dataset which consists of 1,272 (image, description, bounding box) triplets. There are various GUI domains in multiple device environments (Mobile, Desktop, Web) and UI element types (Text, Icon). 

\item \textbf{\textsc{ScreenSpot-v2}} \citet{wu2024atlas} refined the original ScreenSpot dataset by resolving ambiguous descriptions and aligning annotations more precisely with visual elements. We use both versions for a more accurate grounding evaluation.

\item \textbf{\textsc{ScreenSpot-pro}} ScreenSpot-Pro~\citep{li2025screenspotpro} is a benchmark designed to evaluate GUI grounding models in professional, high-resolution environments. It spans 23 applications across five professional categories and three operating systems, highlighting the challenges models face when interacting with complex software..

\item \textbf{Android Control} Android Control~\citep {li2024androidcontrol} is an offline agent benchmark comprising 15000 distinct user tasks sampled from 833 different Android applications. Each task contains a sequence of screenshots, detailed action records, and accessibility trees derived from human demonstrations (the “golden trajectory”). Each action is also annotated with a specific instruction (e.g., “set the hours to 6”), allowing both high-level and low-level task settings. In our experiments, we followed \citet{gou2024uground} and evaluated models on 500 randomly sampled steps from the test split.

\item \textbf{Multimodal-Mind2Web} The Multimodal-Mind2Web~\citep{deng2023mindweb} dataset contains screenshots with a large vertical dimension (e.g., 1280 x 10000 pixels). The test split includes 1,013 realistic user tasks collected from over 100 different websites, each accompanied by a high-level instruction and a sequence of action steps. To make the data manageable we followed \citep{gou2024uground}, and divide the image vertically with certain amount of duplication. During the agent evaluation, when an agent cannot find an actionable element or explicitly chooses to scroll, the next block is processed, simulating user scrolling.

\end{itemize}
\subsection{Model Details}
\begin{itemize}[itemsep=0.05in,leftmargin=0.2in,topsep=0.02in]
\item \textbf{Qwen-2.5-VL} We use Qwen-2.5-VL~\citep{bai2025qwen25vl} is a multimodal large language model (mllm) which inputs text and images and outputs text. The model is equipped with vision-language fusion modules and trained on diverse visual instruction datasets. We trained \sname from Qwen-2.5-VL-3B and 7B models. 

\item \textbf{UGround} UGround~\citep{gou2024uground} is the model trained from Qwen-2-VL~\citep{bai2025qwen25vl}, there are two versions UGround-2B and UGround-7B. They are trained with a synthesized dataset that has 10M images, instructions, and target coordinates triplets. This amount of dataset size is about 500 times bigger than the \sname trained. For the fair comparison, we trained UGround on Qwen-2.5-VL~\citep{bai2025qwen25vl} with the same subset used to train \sname. As shown in Table~\ref{tab:main}, \sname excessively outperforms UGround for both.

\item \textbf{Proprietary Models Endpoint} 
\begin{itemize}[itemsep=0.05in,leftmargin=0.2in,topsep=0.02in]
\item \textit{gpt-4o-2024-05-13}
\item \textit{claude-3-7-sonnet-20250219}
\end{itemize}

\end{itemize}
\subsection{Training and Evaluation Details}\label{appendix:training_setup}
\begin{itemize}[itemsep=0.05in,leftmargin=0.2in,topsep=0.02in]
\item\textbf{Training framework} For the training framework, we utilize VERL for reinforcement learning and TRL for fine-tuning.

\begin{itemize}[itemsep=0.05in,leftmargin=0.2in,topsep=0.02in]
\item TRL (https://github.com/huggingface/trl)
\item VERL (https://github.com/volcengine/verl)
\end{itemize}

\item\textbf{Training hyperparameters.} We describe our training hyperparameters, which we used for training, especially for learning to explaining GUI images via reasoning and learning to predict consistent coordinates under transformations.
\input{resource/tab_hyperparameter}

\end{itemize}

\subsection{Compute Resource}\label{appendix:compute_resource}
For the main development, we mainly use Intel(R) Xeon(R) Gold 6338 CPU @ 2.00GHz and four A100 80GB GPUs. For training \sname, it took around 20 hours.

\clearpage

\section{Additional Experimental Results}

\subsection{Comparison with additional Baselines}\label{appendix:additional_baselines}

\input{resource/tab_morebase}

We compare the grounding accuracy of \sname with other additional baselines, which are closed-source (UI-TARS~\citep{qin2025uitars}) and the concurrently proposed reinforcement learning methods for grounding UI-R1~\citep{lu2025uir1} and GUI-R1~\citep{xia2025guir1}. As shown in Table~\ref{tab:morebase} \sname achieves the highest average combined accuracy (75.3\%) across \textsc{ScreenSpot}~\citep{cheng2024screenspot}, \textsc{ScreenSpot-V2}~\citep{wu2024atlas}, and \textsc{ScreenSpot-Pro}~\citep{li2025screenspotpro}.

Crucially, the compact ReGUIDE-3B already surpasses the best 7B-parameter base model (UI-TARS-7B) by +1.5 pp in average accuracy (72.3 \% $\rightarrow$ 73.8 \%), demonstrating that our training and search strategy substantially improves the GUI grounding performance. Moreover, both reinforcement fine-tuned baselines (i.e., UI-R1-E-3B, GUI-R1-7B) perform worse than \sname, underscoring the effectiveness of our further training for consistency under transformation and search algorithm. The full-sized ReGUIDE-7B further extends this lead to +1.5 pp, establishing a best among the given baselines on the combined benchmarks.

\subsection{Effectiveness of Kernel Density Estimation as majority voting strategy} 

\input{resource/tab_voting}

To verify the most effective method for aggregating diverse pixel-wise coordinate predictions, we conducted an ablation study comparing our Kernel Density Estimation (KDE) based strategy against "Center" and "Medoid". "Center" is simply getting the mean point of the predicted coordinates. "Medoid" selects an actual predicted point that minimizes the sum of distances to all other predictions.  As shown in Table~\ref{tab:ablation_voting}, KDE shows notable results compared with other voting strategies. 

The performance gap can be explained by the statistical properties of each strategy.  
First, the "Center" is vulnerable to outliers, a few erroneous predictions can shift the center far from the true target. Also, "Medoid" is robust to outliers but ignores local density, so it may choose a point in a sparse region when the prediction distribution is multi-modal.  
However, KDE can overcome these limitations. KDE estimating the underlying likelihood surface, naturally capturing local density peaks while down-weighting isolated outliers. This behaviour aligns with our empirical observation that model confidence around the true GUI element is approximately Gaussian (Figure~\ref{figure:confidence}), making KDE a principled and effective spatial voting rule. 

\clearpage
\subsection{Effect of consistency finetuning after reinforcement stage} 
\input{resource/tab_augment}
We assess the benefit of our \emph{consistency-under-transformation} finetuning on the ReGUIDE-3B model using the ScreenSpot benchmark. As shown in Table~\ref{tab:augment}, training GRPO on the original screenshots only (RL w/o Aug.) achieves 83.3 \% accuracy, while mixing the same augmented views directly into the RL phase (RL w/ Aug) yields no improvement (83.2 \%). In contrast, our two-stage pipeline—first training GRPO without augmentation and then applying a separate consistency finetune on the augmented views (\sname)—raises accuracy to 84.9 \%, a gain of +1.6 pp over the vanilla RL model and +1.7 pp over RL w/ Aug. This result confirms that the performance boost arises from explicitly learning to align predictions across transformed views, not from data augmentation alone.

\subsection{Effect of consistency finetuning on transformer attention} 
\input{resource/fig_attention}

To assess how the consistency-under-transformation stage alters internal representations, we visualised the average attention logits of the last transformer layer in \sname-3B. We first removed turn-delimiter and <eos> tokens whose weights dominate the map, and then averaged across all heads. As shown in Figure~\ref{figure:attention}, the GRPO-only model weights most of its attention onto the reasoning span, indicating limited contextual grounding. In contrast, the full \sname model, after consistency finetuning, distributes attention more evenly across reasoning (i.e., text tokens between <think> and </think>) and other text tokens, suggesting richer spatial awareness and reduced attention collapse.

\subsection{Dataset Scalable Experiment} 
\input{resource/fig_data_scale}
To measure how our reinforcement-learning stage benefits from more data, we varied the amount of UGround~\citep{gou2024uground} training samples from 3.2K up to 20K in 3.2K increments and evaluated grounding accuracy (\%) on the \textsc{ScreenSpot}~\citep{cheng2024screenspot} benchmark (Figure~\ref{figure:data_scale}). Accuracy rises monotonically with data size, and the full 20K subset yields the best performance. These results indicate that the \sname training pipeline scales effectively with additional data and has not yet saturated at the 20K level.

\section{Illustration of Searching Strategy.}\label{appendix:searching_strategy}

\input{resource/fig_inference_method}

\section{Alternative RL policy optimization algorithms} \label{appendix:other_rl}
\input{resource/tab_rl_algorithm}
While we primarily adopt GRPO~\citep{shao2024grpo} for the learning to visual reasoning, alternative RL algorithms, such as RLOO~\citep{ahmadian2024rloo}, REINFORCE++~\citep{hu2025reinforce++}, could also be considered. To verify the performance difference between RL poly optimization algorithms, we compared grounding accuracy (\%) on \textsc{ScreenSpot}. As shown in the Table~\ref{tab:rl_algorithm}, GRPO achieves the highest accuracy (83.3\%), followed by RLOO (77.8\%), while REINFORCE++ significantly underperforms (64.8\%). Interestingly, we observed that the REINFORCE++ trained model loses its reasoning ability (i.e., shortest output length), eventually failing to generate meaningful reasoning paths.

\clearpage

\section{Learning Curve} 
We report the learning curve of \sname during training GRPO. The base model of the training curve is Qwen-2.5-VL-3B. We can see that the reward is kept increasing during the training. Also, the mean response length has decreased during the training.
\input{resource/fig_learningcurve}

\section{Prompts}
We describe our generation prompt for \sname (in Table~\ref{tab:ours_prompt} and UGround~\citep{gou2024uground} (in Table~\ref{tab:uground_prompt}. We used the \sname prompt for training and evaluation \sname, and the UGround prompt for supervised fine-tuning the Qwen-2.5-VL for comparison and evaluating the UGround model.

\input{resource/tab_prompt}
\clearpage
\section{Generation Examples}
\input{resource/tab_moregen}

\section{Societal Impact} This paper presents \sname, a method that significantly improves GUI-grounding accuracy. We expect that our approach will enhance digital accessibility and productivity by enabling agents to locate and interact with on-screen elements more reliably. However, the same capability could be misused for malicious purposes, such as automated hacking or spreading misinformation, issues that current agent models already face. Mitigation measures include responsible release practices, access controls, and detection tools to discourage adversarial deployments.

%% file: resource/tab_hyperparameter.tex
\begin{table}[h]
  \centering
  \caption{Hyperparameters for \sname on grpo training}
  \label{tab:hyperparameters_rl}
  \vspace{0.05in}
  \begin{tabular}{l | l}
    \toprule
    \textbf{Hyperparmeter} & \textbf{Value} \\
    \midrule
    Optimizer & Adam\\
    Algorithm & GRPO\\
    Learning rate & 1e-6\\
    Training data size & 20k\\
    Batch size & 128\\
    Generation per Sample & 8\\
    kl\_coef & 0.01\\
    \bottomrule
    \end{tabular}
  \vspace{-0.15in}
\end{table}

\begin{table}[h]
  \centering
  \caption{Hyperparameters for \sname on training consistent coordinated under transformations}
  \label{tab:hyperparameters_sft}
  \vspace{0.05in}
  \begin{tabular}{l | l}
    \toprule
    \textbf{Hyperparmeter} & \textbf{Value} \\
    \midrule
    Optimizer & Adam\\
    Learning rate & 1e-6\\
    Training data size & 20k\\
    Epoch & 1 \\
    Batch size & 64 \\
    minimum crop ratio & 0.3 \\
    \bottomrule
    \end{tabular}
  \vspace{-0.15in}
\end{table}

%% file: resource/tab_morebase.tex
\begin{table*}[h] 
\caption{Accuracy (\%) for \sname (Ours) and other baselines, including UI-TARS~\citep{qin2025uitars}, UI-R1~\citep{lu2025uir1}, GUI-R1~\citep{xia2025guir1}, and \sname (Ours). We evaluate on three web-grounding benchmarks: \textsc{ScreenSpot}~\citep{cheng2024screenspot}, \textsc{ScreenSpot-v2}~\citep{wu2024atlas}, and \textsc{ScreenSpot-Pro}~\citep{li2025screenspotpro}.}
\label{tab:morebase}
\small
\centering
\resizebox{\textwidth}{!}{
\begin{tabular}{lccc|c}
\toprule
Methods & \textsc{ScreenSpot}~\citep{cheng2024screenspot} & \textsc{ScreenSpot-v2}~\citep{wu2024atlas}& \textsc{ScreenSpot-Pro}~\citep{li2025screenspotpro} & Average \\
\midrule
UI-TARS-2B \citep{qin2025uitars} & 82.3 & 84.7& 27.7 & 64.9 \\
UI-TARS-7B \citep{qin2025uitars} & 89.5 & 91.6& 35.7 & 72.3 \\
UI-R1-E-3B \citep{lu2025uir1} & 89.2  & 89.5 & 33.5 & 70.7 \\
GUI-R1-7B \citep{xia2025guir1} & - & - & 31.3 & - \\
\rowcolor{RowHighlight}\textbf{\sname-3B (Ours)} & 87.3 & 89.7& 44.3 & 73.8 \\
\rowcolor{RowHighlight}\textbf{\sname-7B (Ours)}& 90.1 & 91.5& 44.4 & 75.3 \\
\arrayrulecolor{black}\bottomrule
\end{tabular}
}
\vspace{-.1in}
\end{table*}

%% file: resource/tab_voting.tex
\begin{table}[h]
\caption{Comparison of voting algorithm with grounding accuracy (\%) on \textsc{ScreenSpot}~\citep{cheng2024screenspot} and \textsc{ScreenSpot-Pro}~\citep{li2025screenspotpro}} 
\label{tab:ablation_voting}
\centering
\begin{tabular}{l cc}\\\toprule  
Methods & \textsc{ScreenSpot}~\citep{cheng2024screenspot} & \textsc{ScreenSpot-Pro}~\citep{li2025screenspotpro}\\ 
\midrule

Center & 80.3 & 20.2\\ 
Medoid & 84.6 & 29.4 \\  
KDE (Ours) & \textbf{87.3} & \textbf{44.3} \\  
\bottomrule
\end{tabular}
\end{table}

%% file: resource/tab_augment.tex
\begin{table}[h]
\centering
\vspace{-0.1in}
\caption{Effect of consistency‐under‐transformation finetuning on ReGUIDE-3B, evaluated on the 
\vspace{0.05in}
\textsc{ScreenSpot}~\citep{cheng2024screenspot} benchmark.}
\label{tab:augment}
\begin{tabular}{lcc}
\toprule
\textbf{Method} & \textbf{Augmentation usage} & \textbf{Accuracy (\%)} \\
\midrule
RL w/o Aug.     & --  & 83.3 \\
RL w/ Aug.     & In RL stage & 83.2 \\
\sname (Ours).  &  fine-tune   & \textbf{84.9} \\
\bottomrule
\end{tabular}
\end{table}

%% file: resource/fig_attention.tex
\begin{figure*}[h]
\centering
    \begin{subfigure}{0.45\textwidth}
    \centering
    \includegraphics[width=\textwidth]{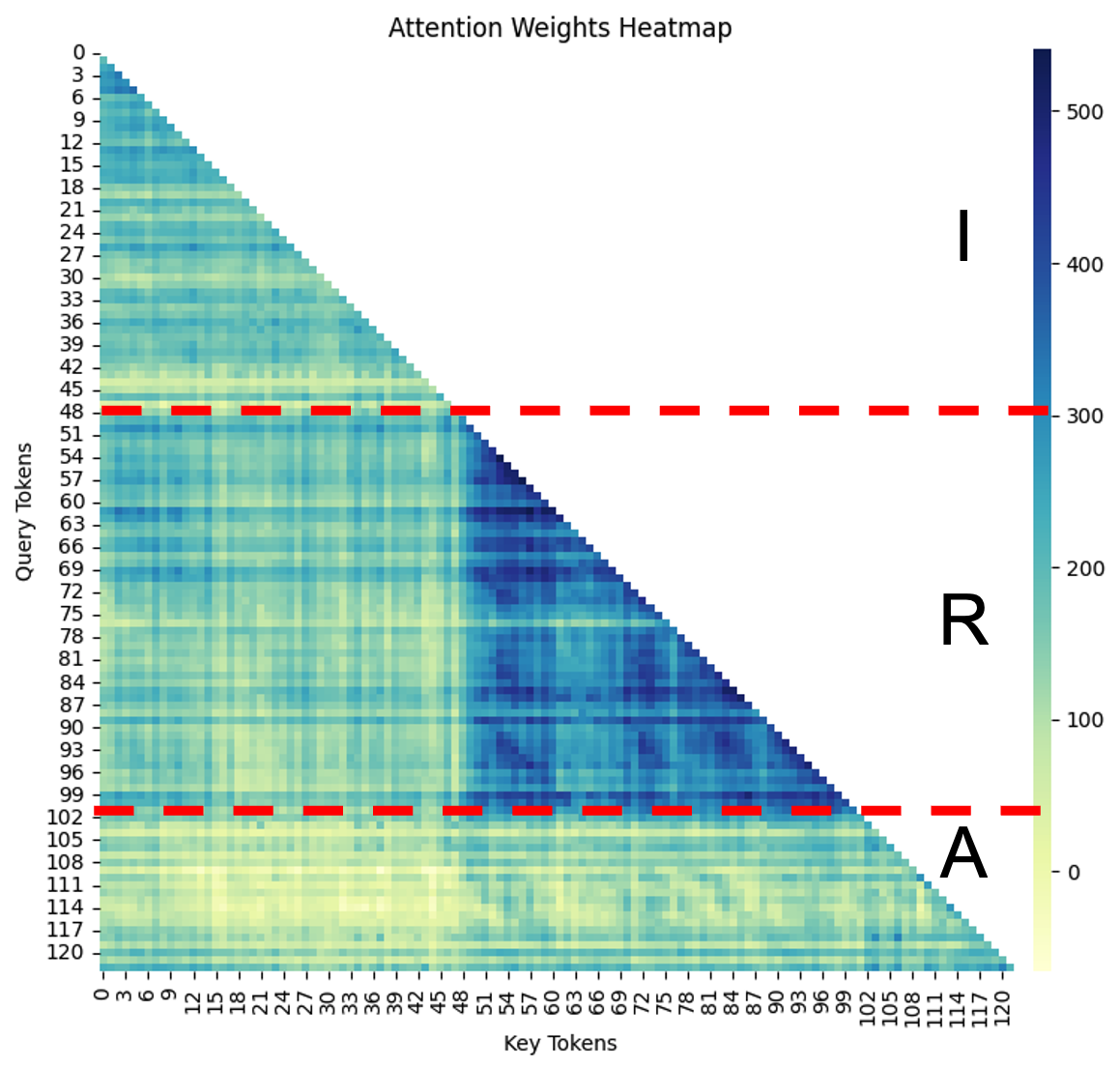}
    \caption{GRPO trained}
    \end{subfigure}
    \begin{subfigure}{0.45\textwidth}
    \centering
    \includegraphics[width=\textwidth]{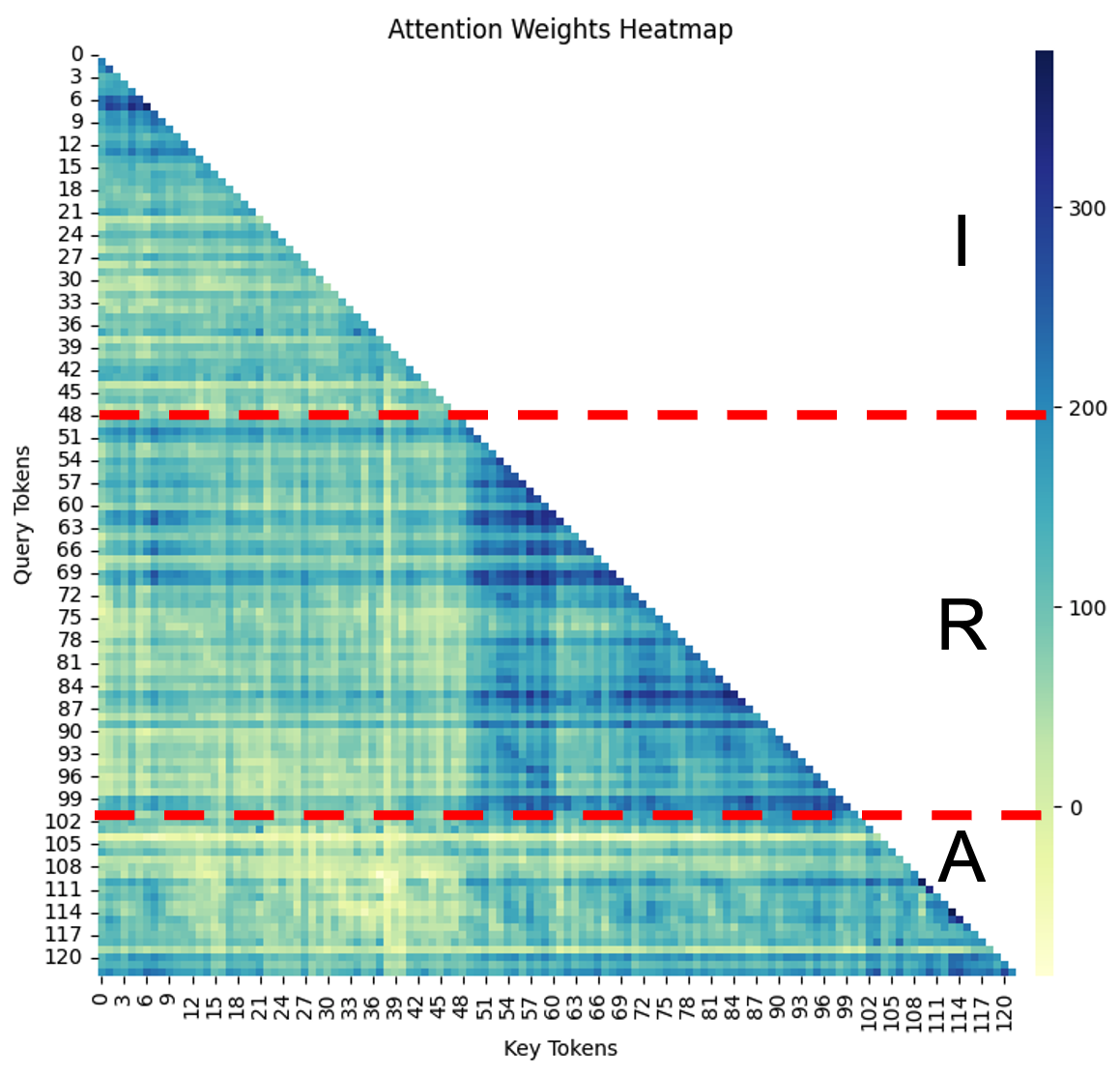}
    \caption{\sname}
    \end{subfigure}
    \caption{Attention map between text tokens, X-axis is key tokens, Y-axis is query tokens. In the image, \textit{I} denotes instruction, \textit{R} denotes reasoning rationales, and \textit{A} denotes Answer.}
    \label{figure:attention}
\end{figure*}

%% file: resource/fig_data_scale.tex
\begin{figure*}[h]
\vspace{-0.1in}
    \centering
    \includegraphics[width=0.4\textwidth]{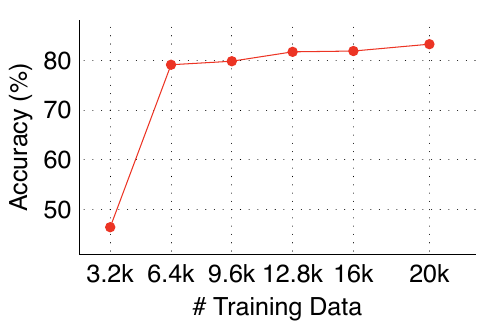}
    
\caption{Grounding accuracy (\%) on \textsc{ScreenSpot}~\citep{cheng2024screenspot} as the UGround~\citep{gou2024uground} training set grows from 3.2 K to 20 K samples. We trained from Qwen-2.5-VL-3B.}
\label{figure:data_scale}
\end{figure*}

%% file: resource/fig_inference_method.tex
\begin{figure*}[h]
\vspace{-0.1in}
    \includegraphics[width=\textwidth]{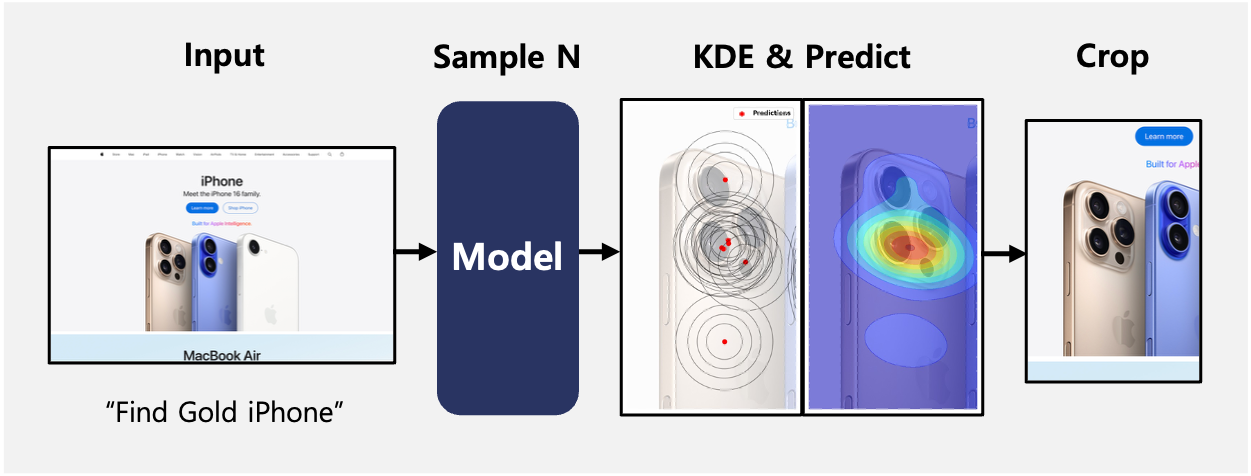}
    \label{figure:inference_concept}
\vspace{-0.1in}
\caption{\textbf{Overview of \sname's searching strategy.}. Starting from the full screenshot, the model samples $N$ coordinate predictions, votes using kernel density estimation, and recenters a crop on the density peak. One searching pass is then run inside the crop, and the point with the highest KDE density is returned as the final prediction.}
\label{figure:method}
\end{figure*}

%% file: resource/tab_rl_algorithm.tex
\begin{table}[h]
\caption{Comparison of RL algorithms. We report accuracy (\%) and average output length (Output Len.) on ScreenSpot Benchmark~\citep{cheng2024screenspot}.} 
\label{tab:rl_algorithm}
\centering
\begin{tabular}{l cc}\\\toprule  
Methods & Accuracy (\%) & Output Len. \\ 
\midrule
REINFORCE++~\citep{hu2025reinforce++} & 64.8 & 32.3 \\ 
RLOO~\citep{ahmadian2024rloo} & 77.8 & 95.7\\  
GRPO~\citep{shao2024grpo} & 83.3 & 80.8 \\  
\bottomrule
\end{tabular}
\end{table}

%% file: resource/fig_learningcurve.tex
\begin{figure*}[h]
\begin{subfigure}{0.47\textwidth}
    \includegraphics[width=\textwidth]{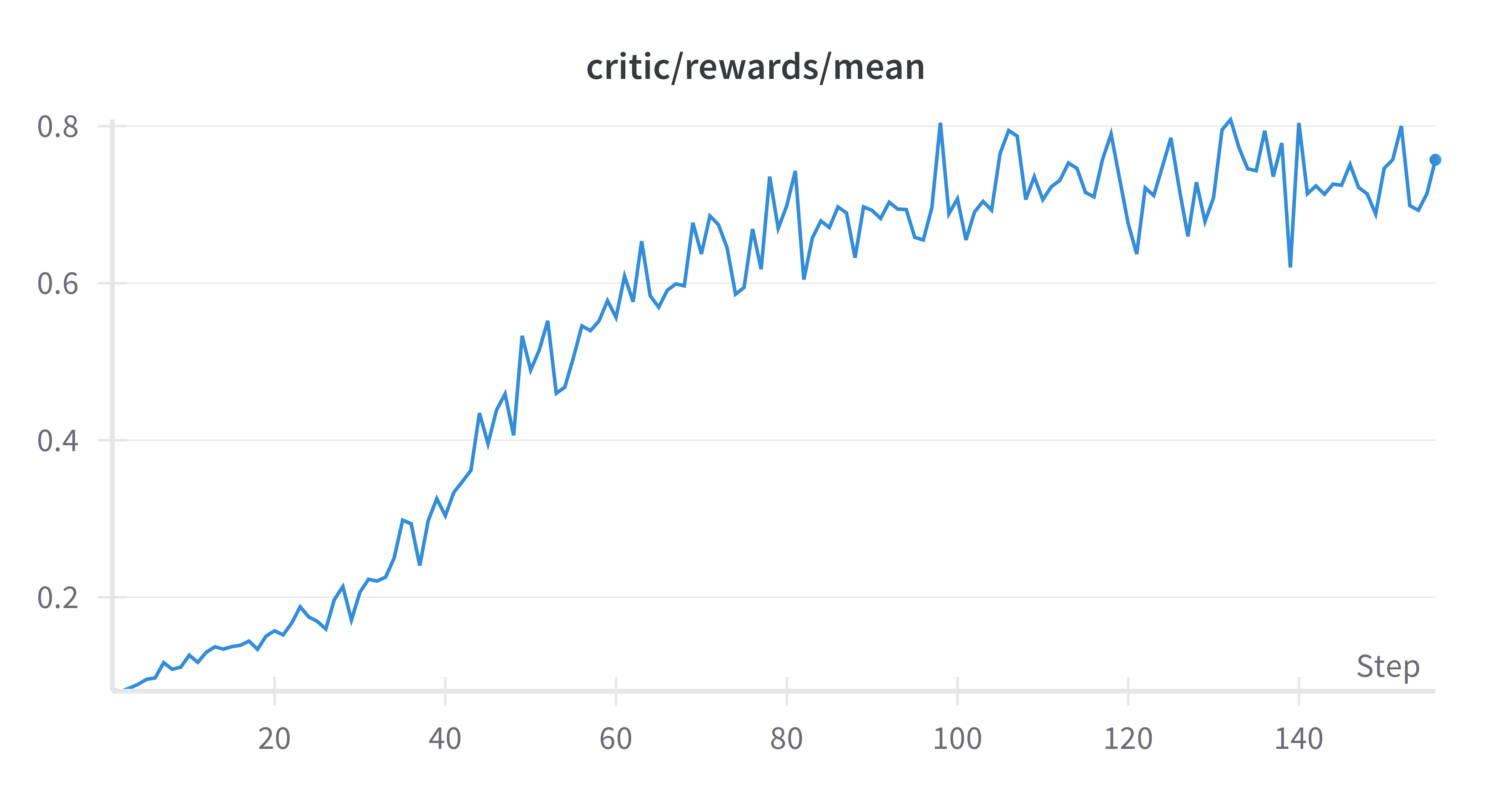}
    \caption{Mean reward score per step}
    \label{figure:mean_reward}
\end{subfigure}
\begin{subfigure}{0.47\textwidth}
    \includegraphics[width=\textwidth]{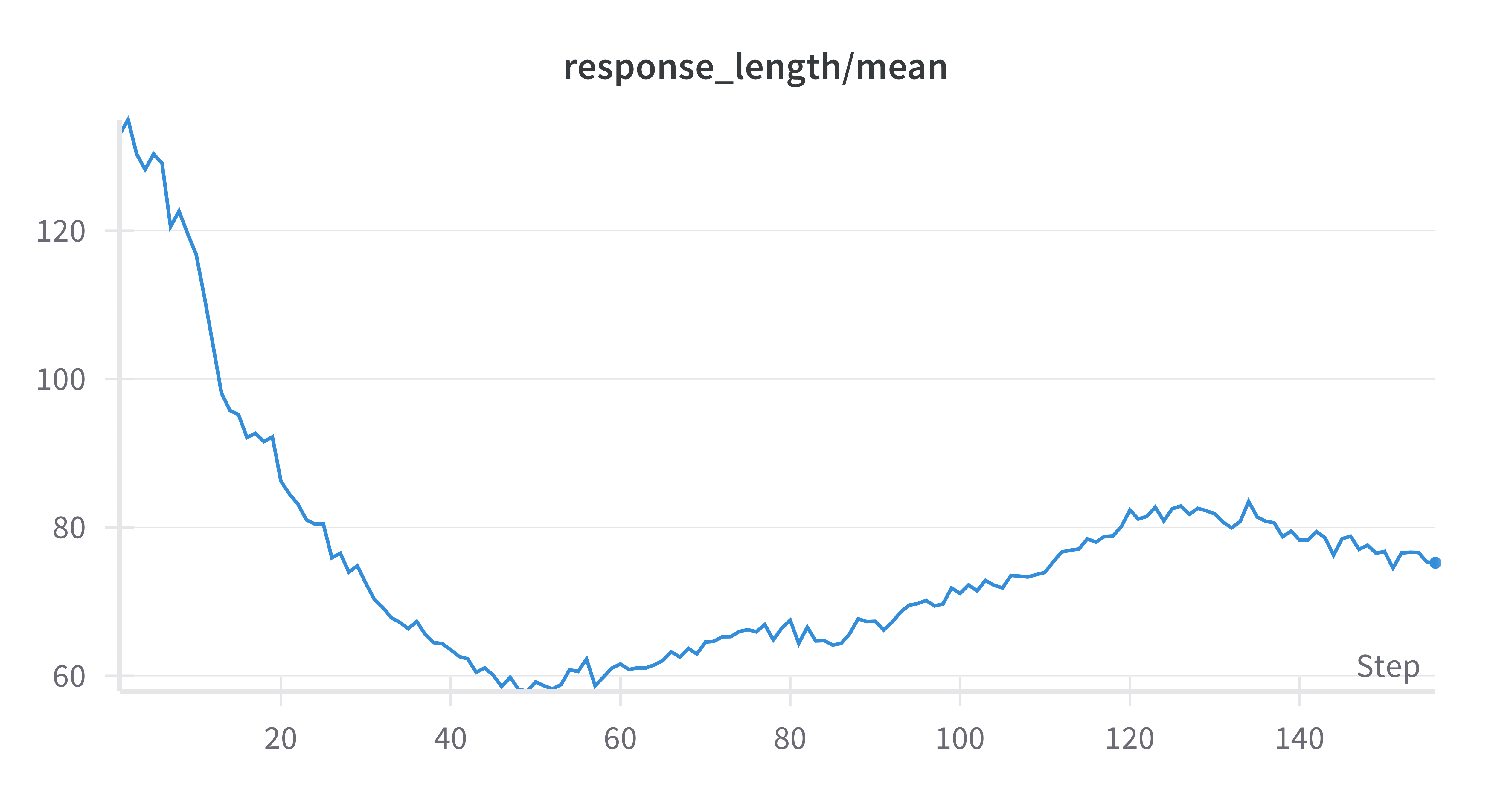}
    \caption{Mean response length per step}
    \label{figure:mean_responselength}
\end{subfigure}
\caption{Learning curve during training GRPO of \sname.}
\label{figure:learning_curve}
\end{figure*}

%% file: resource/tab_prompt.tex
\begin{table}[h]
  \centering
  \caption{Prompt of \sname}
  \label{tab:ours_prompt}
  \begin{tabular}{l}
    \toprule
    \begin{minipage}[t]{\linewidth}
        \textit{Element}: Target element instruction.\\
        \textit{size\_x}: Width of the given image.\\
        \textit{size\_y}: Height of the given image.
    \end{minipage}\\

    \midrule
    \begin{minipage}[t]{\linewidth}
    What is the coordinate of [\{\textit{Element}\}] in the image?\\
    The size of image is (\textit{\{size\_x\}},\textit{\{size\_y\}}).\\
    Output the thinking process in <think> </think> and final answer (coordinate (x,y)) in <answer> </answer> tags.
    \end{minipage}\\
    \bottomrule
    \end{tabular}
  \vspace{-0.15in}
\end{table}

\begin{table}[h]
  \centering
  \caption{Prompt of UGround}
  \label{tab:uground_prompt}
  \begin{tabular}{l}
    \toprule
    \textit{Element}: Target element instruction.\\
    \midrule
    \begin{minipage}[t]{\linewidth}
      Your task is to help the user identify the precise coordinates (x, y) of a specific area/element/object on the screen based on a description.\\
    \\
      - Your response should aim to point to the center or a representative point within the described area/element/object as accurately as possible.\\
      - If the description is unclear or ambiguous, infer the most relevant area or element based on its likely context or purpose.\\
      - Your answer should be a single string (x, y) corresponding to the point of the interest.\\
    \\
      Description: \{\textit{Element}\}\\
    \\
      Answer:
      \end{minipage}\\
    \bottomrule
  \end{tabular}
  \vspace{-0.15in}
\end{table}

%% file: resource/tab_moregen.tex
\begin{table}[h]
  \centering
  \caption{Examples of \sname{}’s generated reasoning and predicted coordinate.}
  \label{tab:more_gen}
  \begin{tabular}{c}
    \toprule
    \raisebox{-0.87\height}{\includegraphics[height=1in]{images/Example4.jpg}}
    \\\\
    \begin{minipage}[t]{\linewidth}
      \textbf{Target Prompt:} add description.\\[0.4em]
      \textbf{Response:} \texttt{<think>}To find the coordinate of [add description], I need to locate the element labeled \"Add description\" in the image. The coordinates are given as (x, y) pairs, and I need to identify the position of this element.\texttt{</think><answer>(300,2221) </answer>}
    \end{minipage} \\
        \midrule
    \raisebox{-0.87\height}{\includegraphics[width=0.33\textwidth]{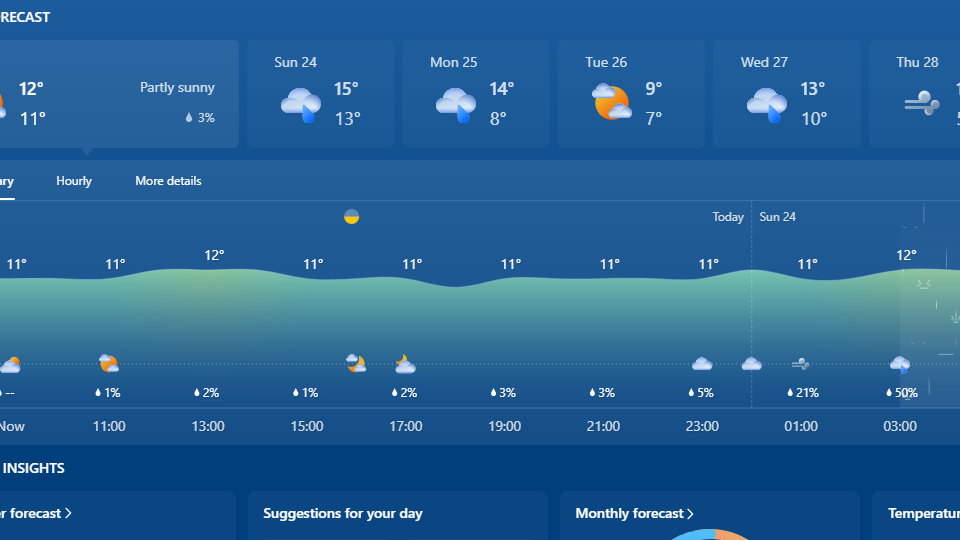}}
    \\\\
    \begin{minipage}[t]{\linewidth}
      \textbf{Target Prompt:} check weather on tuesday.\\[0.4em]
      \textbf{Response:}  \texttt{<think>} To determine the coordinate of the \"nlp\" folder, I need to locate it in the image and identify its position relative to the other folders. The \"nlp\" folder is the second folder in the top row, as indicated by its position in the grid layout. \texttt{</think> <answer>(584,760)</answer>}
    \end{minipage} \\
    \midrule
    \raisebox{-0.87\height}{\includegraphics[width=0.33\textwidth]{images/Example2.jpg}}
    \\\\
    \begin{minipage}[t]{\linewidth}
      \textbf{Target Prompt:} sign up for newsletter.\\[0.4em]
      \textbf{Response:}  \texttt{<think>} The task is to find the coordinate of the \"Sign Up for Newsletter\" checkbox in the image. The checkbox is located in the \"Personal Information\" section, and it is a small square with a label next to it. The coordinates are given in a specific size of the image, and we need to identify the position of the checkbox within the image. \texttt{</think> <answer>(367,85)</answer>}
    \end{minipage} \\
    \midrule
    \raisebox{-0.87\height}{\includegraphics[width=0.33\textwidth]{images/Example3.jpg}}
    \\\\
    \begin{minipage}[t]{\linewidth}
      \textbf{Target Prompt:} start mapping.\\[0.4em]
      \textbf{Response:} \texttt{<think>} The \"Start Mapping\" button is typically used to begin the process of adding or editing data on the map. In the image, it is located in the lower left section, which is a common placement for such buttons on web interfaces. \texttt{</think> <answer>(558,805)</answer>}
    \end{minipage}
    \\
    \bottomrule
  \end{tabular}
  \vspace{-0.15in}
\end{table}

%% file: main.bbl
\begin{thebibliography}{41}
\providecommand{\natexlab}[1]{#1}
\providecommand{\url}[1]{\texttt{#1}}
\expandafter\ifx\csname urlstyle\endcsname\relax
  \providecommand{\doi}[1]{doi: #1}\else
  \providecommand{\doi}{doi: \begingroup \urlstyle{rm}\Url}\fi

\bibitem[Ahmadian et~al.(2024)Ahmadian, Cremer, Gall{\'e}, Fadaee, Kreutzer, Pietquin, {\"U}st{\"u}n, and Hooker]{ahmadian2024rloo}
A.~Ahmadian, C.~Cremer, M.~Gall{\'e}, M.~Fadaee, J.~Kreutzer, O.~Pietquin, A.~{\"U}st{\"u}n, and S.~Hooker.
\newblock Back to basics: Revisiting reinforce style optimization for learning from human feedback in llms.
\newblock \emph{arXiv preprint arXiv:2402.14740}, 2024.

\bibitem[Alibaba(2025)]{bai2025qwen25vl}
Alibaba.
\newblock Qwen2.5-vl technical report.
\newblock \emph{arXiv preprint arXiv:abs/2502.13923}, 2025.

\bibitem[Anthropic.(2024)]{anthropic2024computeruse}
T.~Anthropic.
\newblock Introducing computer use, a new claude 3.5 sonnet, and claude 3.5 haiku.
\newblock \url{https://www.anthropic.com/news/3-5-models-and-computer-use}, October 2024.

\bibitem[Anthropic.(2025)]{anthropic2024claude}
T.~Anthropic.
\newblock Claude 3.7 sonnet and claude code.
\newblock \url{https://www.anthropic.com/news/claude-3-7-sonnet}, February 2025.

\bibitem[Caron et~al.(2021)Caron, Touvron, Misra, J{\'e}gou, Mairal, Bojanowski, and Joulin]{caron2021dino}
M.~Caron, H.~Touvron, I.~Misra, H.~J{\'e}gou, J.~Mairal, P.~Bojanowski, and A.~Joulin.
\newblock Emerging properties in self-supervised vision transformers.
\newblock In \emph{Proceedings of the IEEE/CVF international conference on computer vision}, pages 9650--9660, 2021.

\bibitem[Chen et~al.(2025)Chen, Zhang, Khalilov, Guo, Gebreegziabher, Ye, Xiao, Yao, Li, and Li]{chen2025toward}
C.~Chen, Z.~Zhang, I.~Khalilov, B.~Guo, S.~A. Gebreegziabher, Y.~Ye, Z.~Xiao, Y.~Yao, T.~Li, and T.~J.-J. Li.
\newblock Toward a human-centered evaluation framework for trustworthy llm-powered gui agents.
\newblock \emph{arXiv preprint arXiv:2504.17934}, 2025.

\bibitem[Cheng et~al.(2024{\natexlab{a}})Cheng, Sun, Chu, Xu, Li, Zhang, and Wu]{cheng2024screenspot}
K.~Cheng, Q.~Sun, Y.~Chu, F.~Xu, Y.~Li, J.~Zhang, and Z.~Wu.
\newblock Seeclick: Harnessing gui grounding for advanced visual gui agents.
\newblock \emph{arXiv preprint arXiv:2401.10935}, 2024{\natexlab{a}}.

\bibitem[Cheng et~al.(2024{\natexlab{b}})Cheng, Sun, Chu, Xu, Li, Zhang, and Wu]{cheng2024seeclick}
K.~Cheng, Q.~Sun, Y.~Chu, F.~Xu, Y.~Li, J.~Zhang, and Z.~Wu.
\newblock Seeclick: Harnessing gui grounding for advanced visual gui agents.
\newblock \emph{arXiv preprint arXiv:2401.10935}, 2024{\natexlab{b}}.

\bibitem[DeepSeek(2024)]{shao2024grpo}
DeepSeek.
\newblock Deepseekmath: Pushing the limits of mathematical reasoning in open language models.
\newblock \emph{arXiv preprint arXiv:2402.03300}, 2024.

\bibitem[Deng et~al.(2023)Deng, Gu, Zheng, Chen, Stevens, Wang, Sun, and Su]{deng2023mindweb}
X.~Deng, Y.~Gu, B.~Zheng, S.~Chen, S.~Stevens, B.~Wang, H.~Sun, and Y.~Su.
\newblock Mind2web: Towards a generalist agent for the web.
\newblock In \emph{Thirty-seventh Conference on Neural Information Processing Systems}, 2023.
\newblock URL \url{https://openreview.net/forum?id=kiYqbO3wqw}.

\bibitem[Gou et~al.(2024)Gou, Wang, Zheng, Xie, Chang, Shu, Sun, and Su]{gou2024uground}
B.~Gou, R.~Wang, B.~Zheng, Y.~Xie, C.~Chang, Y.~Shu, H.~Sun, and Y.~Su.
\newblock Navigating the digital world as humans do: Universal visual grounding for gui agents.
\newblock \emph{arXiv preprint arXiv:2410.05243}, 2024.

\bibitem[Guo et~al.(2025)Guo, Yang, Zhang, Song, Zhang, Xu, Zhu, Ma, Wang, Bi, et~al.]{guo2025deepseekr1}
D.~Guo, D.~Yang, H.~Zhang, J.~Song, R.~Zhang, R.~Xu, Q.~Zhu, S.~Ma, P.~Wang, X.~Bi, et~al.
\newblock Deepseek-r1: Incentivizing reasoning capability in llms via reinforcement learning.
\newblock \emph{arXiv preprint arXiv:2501.12948}, 2025.

\bibitem[Hjelm et~al.(2018)Hjelm, Fedorov, Lavoie-Marchildon, Grewal, Bachman, Trischler, and Bengio]{hjelm2018learning}
R.~D. Hjelm, A.~Fedorov, S.~Lavoie-Marchildon, K.~Grewal, P.~Bachman, A.~Trischler, and Y.~Bengio.
\newblock Learning deep representations by mutual information estimation and maximization.
\newblock \emph{arXiv preprint arXiv:1808.06670}, 2018.

\bibitem[Hong et~al.(2024)Hong, Wang, Lv, Xu, Yu, Ji, Wang, Wang, Dong, Ding, et~al.]{hong2024cogagent}
W.~Hong, W.~Wang, Q.~Lv, J.~Xu, W.~Yu, J.~Ji, Y.~Wang, Z.~Wang, Y.~Dong, M.~Ding, et~al.
\newblock Cogagent: A visual language model for gui agents.
\newblock In \emph{Proceedings of the IEEE/CVF Conference on Computer Vision and Pattern Recognition}, pages 14281--14290, 2024.

\bibitem[Hosseini et~al.(2024)Hosseini, Yuan, Malkin, Courville, Sordoni, and Agarwal]{hosseini2024vst}
A.~Hosseini, X.~Yuan, N.~Malkin, A.~Courville, A.~Sordoni, and R.~Agarwal.
\newblock V-star: Training verifiers for self-taught reasoners.
\newblock \emph{arXiv preprint arXiv:2402.06457}, 2024.

\bibitem[Hu(2025)]{hu2025reinforce++}
J.~Hu.
\newblock Reinforce++: A simple and efficient approach for aligning large language models.
\newblock \emph{arXiv preprint arXiv:2501.03262}, 2025.

\bibitem[Huang et~al.(2025)Huang, Jia, Zhai, Cao, Ye, Zhao, Xu, Hu, and Lin]{huang2025visionr1}
W.~Huang, B.~Jia, Z.~Zhai, S.~Cao, Z.~Ye, F.~Zhao, Z.~Xu, Y.~Hu, and S.~Lin.
\newblock Vision-r1: Incentivizing reasoning capability in multimodal large language models.
\newblock \emph{arXiv preprint arXiv:2503.06749}, 2025.

\bibitem[Lampinen et~al.(2022)Lampinen, Dasgupta, Chan, Matthewson, Tessler, Creswell, McClelland, Wang, and Hill]{lampinen2022can}
A.~K. Lampinen, I.~Dasgupta, S.~C. Chan, K.~Matthewson, M.~H. Tessler, A.~Creswell, J.~L. McClelland, J.~X. Wang, and F.~Hill.
\newblock Can language models learn from explanations in context?
\newblock \emph{arXiv preprint arXiv:2204.02329}, 2022.

\bibitem[Lee et~al.(2025)Lee, Oh, Kim, Shin, and Tack]{lee2025revise}
H.~Lee, S.~Oh, J.~Kim, J.~Shin, and J.~Tack.
\newblock Revise: Learning to refine at test-time via intrinsic self-verification.
\newblock \emph{arXiv preprint arXiv:2502.14565}, 2025.

\bibitem[Li et~al.(2025)Li, Meng, Lin, Luo, Tian, Ma, Huang, and Chua]{li2025screenspotpro}
K.~Li, Z.~Meng, H.~Lin, Z.~Luo, Y.~Tian, J.~Ma, Z.~Huang, and T.-S. Chua.
\newblock Screenspot-pro: Gui grounding for professional high-resolution computer use.
\newblock \emph{arXiv preprint arXiv:2504.07981}, 2025.

\bibitem[Li et~al.(2024)Li, Bishop, Li, Rawles, Campbell-Ajala, Tyamagundlu, and Riva]{li2024androidcontrol}
W.~Li, W.~Bishop, A.~Li, C.~Rawles, F.~Campbell-Ajala, D.~Tyamagundlu, and O.~Riva.
\newblock On the effects of data scale on computer control agents.
\newblock \emph{arXiv preprint arXiv:2406.03679}, 2024.

\bibitem[Lin et~al.(2024)Lin, Li, Gao, Yang, Wu, Bai, Lei, Wang, and Shou]{lin2024showui}
K.~Q. Lin, L.~Li, D.~Gao, Z.~Yang, S.~Wu, Z.~Bai, W.~Lei, L.~Wang, and M.~Z. Shou.
\newblock Showui: One vision-language-action model for gui visual agent.
\newblock \emph{arXiv preprint arXiv:2411.17465}, 2024.

\bibitem[Lu et~al.(2025)Lu, Chai, Guo, Yin, Liu, Wang, Xiong, and Li]{lu2025uir1}
Z.~Lu, Y.~Chai, Y.~Guo, X.~Yin, L.~Liu, H.~Wang, G.~Xiong, and H.~Li.
\newblock Ui-r1: Enhancing action prediction of gui agents by reinforcement learning.
\newblock \emph{arXiv preprint arXiv:2503.21620}, 2025.

\bibitem[Ma et~al.(2025)Ma, Ding, Luo, Chen, Guo, Wong, Feng, and Sun]{ma2025deepperception}
X.~Ma, Z.~Ding, Z.~Luo, C.~Chen, Z.~Guo, D.~F. Wong, X.~Feng, and M.~Sun.
\newblock Deepperception: Advancing r1-like cognitive visual perception in mllms for knowledge-intensive visual grounding.
\newblock \emph{arXiv preprint arXiv:2503.12797}, 2025.

\bibitem[Nye et~al.(2021)Nye, Andreassen, Gur-Ari, Michalewski, Austin, Bieber, Dohan, Lewkowycz, Bosma, Luan, et~al.]{nye2021show}
M.~Nye, A.~J. Andreassen, G.~Gur-Ari, H.~Michalewski, J.~Austin, D.~Bieber, D.~Dohan, A.~Lewkowycz, M.~Bosma, D.~Luan, et~al.
\newblock Show your work: Scratchpads for intermediate computation with language models.
\newblock \emph{arXiv preprint arXiv:2112.00114}, 2021.

\bibitem[OpenAI(2024)]{openai2024gpt4ocard}
OpenAI.
\newblock Gpt-4o system card.
\newblock \emph{arXiv preprint arXiv:2410.21276}, 2024.

\bibitem[Oquab et~al.(2023)Oquab, Darcet, Moutakanni, Vo, Szafraniec, Khalidov, Fernandez, Haziza, Massa, El-Nouby, et~al.]{oquab2023dinov2}
M.~Oquab, T.~Darcet, T.~Moutakanni, H.~Vo, M.~Szafraniec, V.~Khalidov, P.~Fernandez, D.~Haziza, F.~Massa, A.~El-Nouby, et~al.
\newblock Dinov2: Learning robust visual features without supervision.
\newblock \emph{arXiv preprint arXiv:2304.07193}, 2023.

\bibitem[Qin et~al.(2025)Qin, Ye, Fang, Wang, Liang, Tian, Zhang, Li, Li, Huang, et~al.]{qin2025uitars}
Y.~Qin, Y.~Ye, J.~Fang, H.~Wang, S.~Liang, S.~Tian, J.~Zhang, J.~Li, Y.~Li, S.~Huang, et~al.
\newblock Ui-tars: Pioneering automated gui interaction with native agents.
\newblock \emph{arXiv preprint arXiv:2501.12326}, 2025.

\bibitem[Rennie et~al.(2017)Rennie, Marcheret, Mroueh, Ross, and Goel]{rennie2017selfcriticim}
S.~J. Rennie, E.~Marcheret, Y.~Mroueh, J.~Ross, and V.~Goel.
\newblock Self-critical sequence training for image captioning.
\newblock In \emph{IEEE Conference on Computer Vision and Pattern Recognition}, pages 7008--7024, 2017.

\bibitem[Rosenblat(1956)]{rosenblat1956remarks}
M.~Rosenblat.
\newblock Remarks on some nonparametric estimates of a density function.
\newblock \emph{Ann. Math. Stat}, 27:\penalty0 832--837, 1956.

\bibitem[Shaw et~al.(2023)Shaw, Joshi, Cohan, Berant, Pasupat, Hu, Khandelwal, Lee, and Toutanova]{shaw2023pixels}
P.~Shaw, M.~Joshi, J.~Cohan, J.~Berant, P.~Pasupat, H.~Hu, U.~Khandelwal, K.~Lee, and K.~N. Toutanova.
\newblock From pixels to ui actions: Learning to follow instructions via graphical user interfaces.
\newblock \emph{Advances in Neural Information Processing Systems}, 36:\penalty0 34354--34370, 2023.

\bibitem[Snell et~al.(2024)Snell, Lee, Xu, and Kumar]{snell2024scaling}
C.~Snell, J.~Lee, K.~Xu, and A.~Kumar.
\newblock Scaling llm test-time compute optimally can be more effective than scaling model parameters.
\newblock \emph{arXiv preprint arXiv:2408.03314}, 2024.

\bibitem[Wang et~al.(2024)Wang, Liu, Chen, Zhou, Gan, Zeng, Che, Yu, Hao, Shao, et~al.]{wang2024guisurv}
S.~Wang, W.~Liu, J.~Chen, Y.~Zhou, W.~Gan, X.~Zeng, Y.~Che, S.~Yu, X.~Hao, K.~Shao, et~al.
\newblock Gui agents with foundation models: A comprehensive survey.
\newblock \emph{arXiv preprint arXiv:2411.04890}, 2024.

\bibitem[Wu and Xie(2024)]{wu2024vstar}
P.~Wu and S.~Xie.
\newblock V*: Guided visual search as a core mechanism in multimodal llms.
\newblock In \emph{Proceedings of the IEEE/CVF Conference on Computer Vision and Pattern Recognition}, pages 13084--13094, 2024.

\bibitem[Wu et~al.(2024)Wu, Wu, Xu, Wang, Sun, Jia, Cheng, Ding, Chen, Liang, et~al.]{wu2024atlas}
Z.~Wu, Z.~Wu, F.~Xu, Y.~Wang, Q.~Sun, C.~Jia, K.~Cheng, Z.~Ding, L.~Chen, P.~P. Liang, et~al.
\newblock Os-atlas: A foundation action model for generalist gui agents.
\newblock \emph{arXiv preprint arXiv:2410.23218}, 2024.

\bibitem[Xia and Luo(2025)]{xia2025guir1}
X.~Xia and R.~Luo.
\newblock Gui-r1: A generalist r1-style vision-language action model for gui agents.
\newblock \emph{arXiv preprint arXiv:2504.10458}, 2025.

\bibitem[Xu et~al.(2024)Xu, Wang, Wang, Lu, Xie, Saha, Sahoo, Yu, and Xiong]{xu2024aguvis}
Y.~Xu, Z.~Wang, J.~Wang, D.~Lu, T.~Xie, A.~Saha, D.~Sahoo, T.~Yu, and C.~Xiong.
\newblock Aguvis: Unified pure vision agents for autonomous gui interaction.
\newblock \emph{arXiv preprint arXiv:2412.04454}, 2024.

\bibitem[Yu et~al.(2024)Yu, Yao, Zhang, He, Han, Cui, Hu, Liu, Zheng, Sun, et~al.]{yu2024rlhfv}
T.~Yu, Y.~Yao, H.~Zhang, T.~He, Y.~Han, G.~Cui, J.~Hu, Z.~Liu, H.-T. Zheng, M.~Sun, et~al.
\newblock Rlhf-v: Towards trustworthy mllms via behavior alignment from fine-grained correctional human feedback.
\newblock In \emph{Proceedings of the IEEE/CVF Conference on Computer Vision and Pattern Recognition}, pages 13807--13816, 2024.

\bibitem[Zhang et~al.(2025)Zhang, Ding, Ma, Chen, Sun, Lan, and He]{zhang2025breaking}
J.~Zhang, Z.~Ding, C.~Ma, Z.~Chen, Q.~Sun, Z.~Lan, and J.~He.
\newblock Breaking the data barrier--building gui agents through task generalization.
\newblock \emph{arXiv preprint arXiv:2504.10127}, 2025.

\bibitem[Zhao et~al.(2025)Zhao, Chen, and Wang]{zhao2025robustness}
H.~Zhao, T.~Chen, and Z.~Wang.
\newblock On the robustness of gui grounding models against image attacks.
\newblock \emph{arXiv preprint arXiv:2504.04716}, 2025.

\bibitem[Zheng et~al.(2024)Zheng, Gou, Kil, Sun, and Su]{zheng2024seeact}
B.~Zheng, B.~Gou, J.~Kil, H.~Sun, and Y.~Su.
\newblock Gpt-4v(ision) is a generalist web agent, if grounded.
\newblock In \emph{Forty-first International Conference on Machine Learning}, 2024.
\newblock URL \url{https://openreview.net/forum?id=piecKJ2DlB}.

\end{thebibliography}
